\documentclass[10pt,twocolumn,letterpaper]{article}

\usepackage[pagenumbers]{cvpr}              
\usepackage[accsupp]{axessibility}
\usepackage[dvipsnames]{xcolor}

\usepackage{CJKutf8}

\usepackage{amsmath}

\usepackage{colortbl}
\usepackage{color}
\usepackage[dvipsnames]{xcolor}

\definecolor{Gray}{gray}{0.77}
\definecolor{DarkGray}{gray}{0.43}
\definecolor{LLGray}{gray}{0.91}
\definecolor{LGray}{gray}{0.94}
\definecolor{LightCyan}{rgb}{0.88,1,1}
\definecolor{Greenish}{rgb}{0.10,0.60,0.30}
\definecolor{ForestGreen}{RGB}{34,139,34}
\definecolor{Wood}{RGB}{150,111,51}
\definecolor{DarkBrown}{RGB}{150,78,2}
\definecolor{Pinkish}{RGB}{255,102,220}
\definecolor{Blueish}{RGB}{2,78,150}

\definecolor{cvprblue}{rgb}{0.21,0.49,0.74}

\definecolor{LGray}{gray}{0.94}
\definecolor{LightCyan}{rgb}{0.90,1,1}

\newcolumntype{g}{>{\columncolor{LightCyan}}c}

            \usepackage[pagebackref,breaklinks,colorlinks,citecolor=cvprblue]{hyperref}

\def\paperID{4711} 
\def\confName{CVPR}
\def\confYear{2024}

\title{Incorporating Geo-Diverse Knowledge into Prompting for Increased Geographical Robustness in Object Recognition}

\author{Kyle Buettner\textsuperscript{\rm 1}, Sina Malakouti\textsuperscript{\rm 2}, Xiang Lorraine Li\textsuperscript{\rm 1,2},
    Adriana Kovashka\textsuperscript{\rm 1,2}\\
\textsuperscript{1}Intelligent Systems Program, \textsuperscript{2}Department of Computer Science,
University of Pittsburgh, PA, USA\\
{\tt\small \{buettnerk, sem238\}@pitt.edu, \{xianglli, kovashka\}@cs.pitt.edu} \\
\tt\small \href{https://krbuettner.github.io/GeoKnowledgePrompting}{https://krbuettner.github.io/GeoKnowledgePrompting}
}

\begin{document}
\maketitle
\begin{abstract}


    Existing object recognition models have been shown to lack robustness in diverse geographical scenarios due to domain shifts in design and context. Class representations need to be adapted to more accurately reflect an object concept under these shifts. In the absence of training data from target geographies, we hypothesize that geographically diverse descriptive knowledge of categories can enhance robustness. For this purpose, we explore the feasibility of probing a large language model for geography-based object knowledge, and we examine the effects of integrating knowledge into zero-shot and learnable soft prompting with CLIP. Within this exploration, we propose geography knowledge regularization to ensure that soft prompts trained on a source set of geographies generalize to an unseen target set. Accuracy gains over prompting baselines on DollarStreet while training only on Europe data are up to +2.8/1.2/1.6 on target data from Africa/Asia/Americas, and +4.6 overall on the hardest classes. Competitive performance is shown vs. few-shot target training, and analysis is provided to direct future study of geographical robustness.
\end{abstract}    

\section{Introduction}
\label{sec:intro}

        The performance of object recognition models degrades when tested in new geographies (e.g., cities, countries, continents) \cite{de2019does, shankar2017no,kalluri2023geonet, prabhu2022can,wang2020train}. Numerous factors contribute to the challenging problem of \textit{geographical domain shift}, such as cross-geography changes in object design/parts, materials, and context. 
        These changes in turn may be due to cultural, climate, or economic differences around the world. 
        Recent work has shown standard adaptation techniques fail when used for geographical domain shifts \cite{kalluri2023geonet, prabhu2022can}, but there has yet to be significant progress in the creation of techniques that improve geographical robustness. Such progress is necessary to ensure equitable use of AI in the future.

        \begin{figure}[t]
    \centering
    \includegraphics[scale=0.24]
    {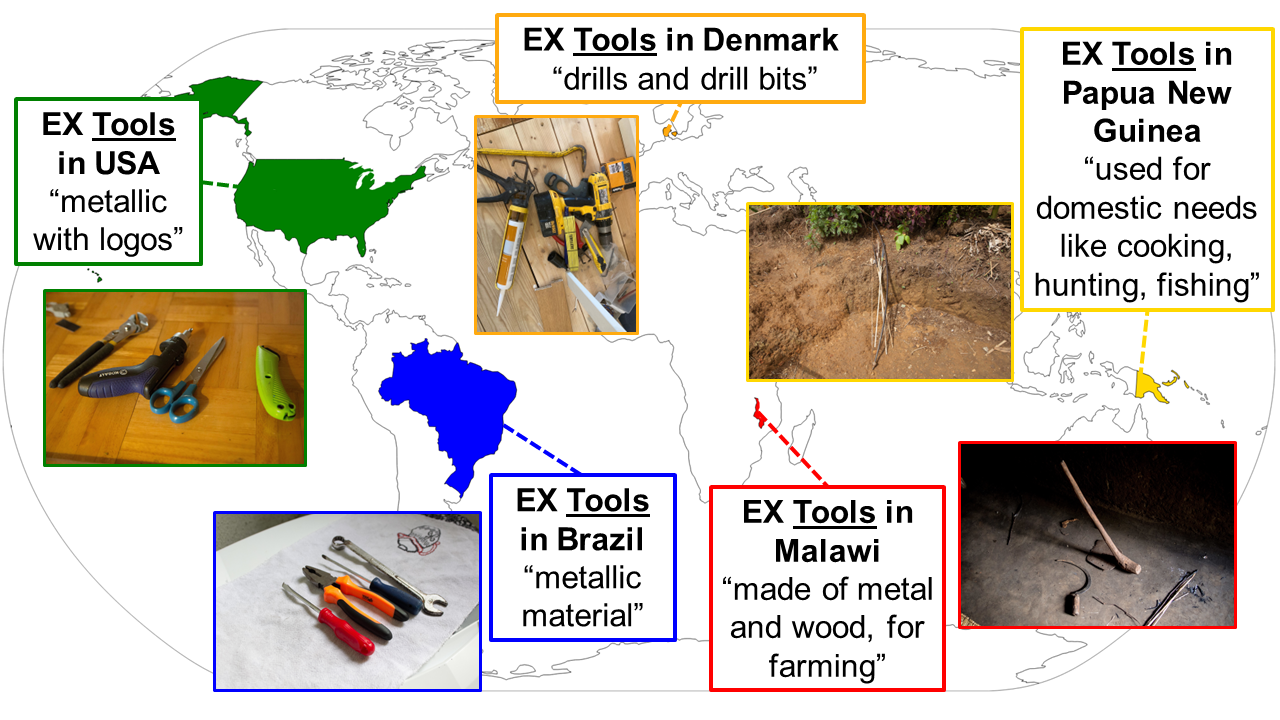}
    \vspace{-3.5mm}
    \caption{\textbf{Descriptive knowledge can address concept shifts across geographies.} Observe the wide range of object designs and contexts in the DollarStreet \cite{NEURIPS2022_5474d9d4} category \textit{tools} around the world. Our work's premise is that textual representations for classes in vision-language models can be enhanced to better suit diverse object representations across geographies. 
    Map made with \cite{plotly}. }
    \label{intro_fig}
\end{figure}

        Overall, models need representations that 
        adequately capture a category's various forms around the world. A natural solution is to collect training data of objects from different regions. However, this approach is expensive, takes significant effort, and is difficult for regions with limited Internet access. Fortunately, geographical shifts have a unique property compared to other common domain shifts (\eg ones due to artistic style or weather changes)---they can be addressed with \emph{descriptive knowledge} about concept changes. In other words, it is possible to describe the features of an object in a region and use this information to adapt a model's default representation. For instance, as shown in Fig.~\ref{intro_fig}, for rural areas in Papua New Guinea, \textit{tools} can be described as being used for ``cooking, hunting, and fishing'', and for rural areas in Malawi, \textit{tools} may often be ``made of metal and wood, for farming''. Models should account for diverse presentations and contexts of a category and not be limited to biased presentations (\eg if the model learns \textit{tools} as just being ``metallic with logos''). 

        We examine the effects of probing geo-diverse knowledge in two ways. 
        First, we analyze whether a vision-language model (VLM, i.e. CLIP \cite{radford2021learning}) has encoded categories in a geo-specific manner, such that adding a country's name to a prompt (\eg ``A photo of a house in China'') elicits knowledge that improves recognition. Second, we probe a large language model (LLM, i.e. GPT-3 \textit{davinci-003}) for geography-specific knowledge to obtain visual feature descriptors for an object in different locations. We analyze results in zero-shot inference on geographically and socioeconomically diverse data (DollarStreet \cite{NEURIPS2022_5474d9d4}), finding the combination of knowledge to often be complementary. 
        
        We further consider a practical scenario where CLIP is optimized with soft prompting, using only a ``source'' geography with easy-to-access data (\eg Europe), while the model is applied downstream on ``target'' data from other parts of the world (\eg Africa, Asia, Americas). We propose \emph{geography knowledge regularization}, which uses knowledge ensembled over countries to enable soft prompts to achieve geographically generalizable class representations. We test our method on the recent DollarStreet and GeoNet \cite{kalluri2023geonet} datasets.
        Our regularization boosts performance over baseline soft prompting methods, and
        has benefits with respect to few-shot target-specific training (a 16-shot-per-class regularized model without any target data outperforms a 12-shot-per-class target-trained model on DollarStreet).

        Our method is the first to effectively address geo shifts in object recognition. It outperforms zero-shot CLIP (assumed to have some robustness) by 10.3\% on Africa, CoOp \cite{zhou2022learning} by 3.3\%, and the best baseline by 4.6\% on the hardest classes. 
       
        To summarize, we answer the following questions:
        (1) Does adding geographical context (i.e. country names) to CLIP prompts improve recognition across geographies?
        (2) Can an LLM provide useful geographical descriptive knowledge to improve recognition? 
        (3) How can we optimize soft prompts for CLIP using an accessible data source with consideration of target geographies not represented in the training set?  
        (4) Where can soft prompts enhanced with geographical knowledge provide the most benefits?

\begin{figure*}[t]
    \centering
    \includegraphics[scale=0.38]
    {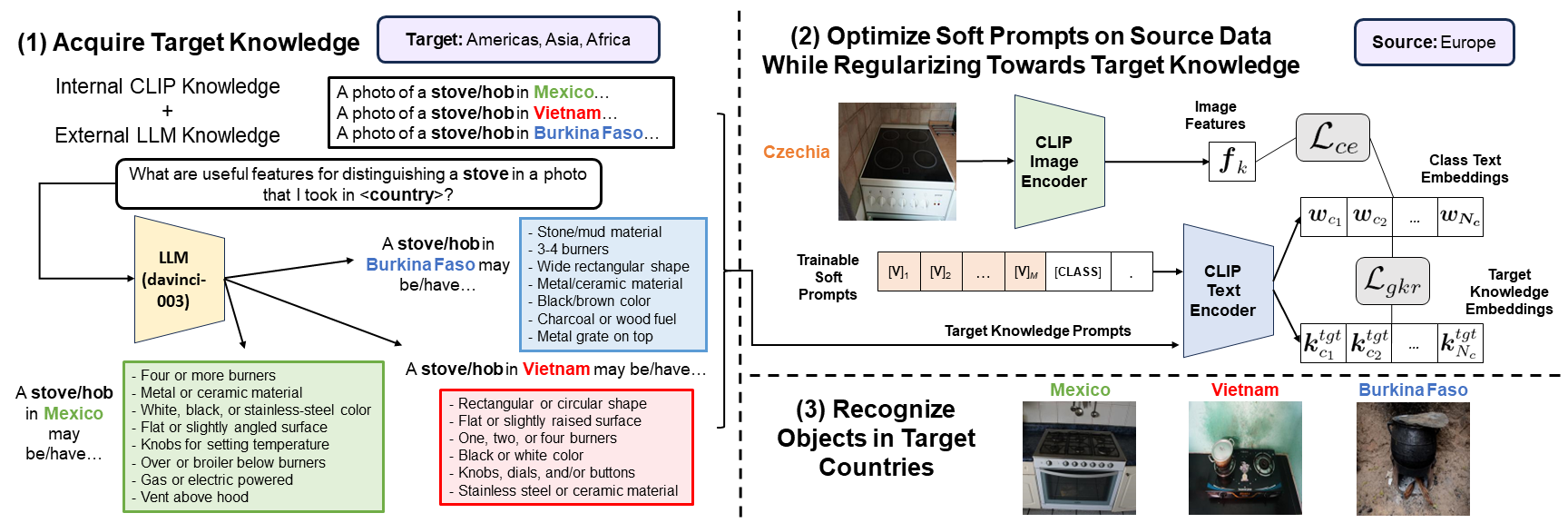}
    \vspace{-3mm}
    \caption{\textbf{Geography knowledge regularization.} To ensure robustness in soft prompt learning, we (1) incorporate knowledge internal to CLIP and externally obtained from an LLM. (2) This descriptive knowledge regularizes class representations when training on a specific source geography (\eg Europe), thus (3) increasing robustness when generalizing to target geographies (\eg Vietnam).}
    \label{approach_fig}
\end{figure*}

\section{Related Work}
    \label{sec:related work}

    \noindent \textbf{Geographical domain shifts} 
        occur when the target setting is in a different geography (\eg continent, country, city) than where the source data was acquired. Shifts involve changes in object design (\eg differences in house architecture) and context (i.e. background/co-occurring objects vary). Datasets tailored to cross-country/continent object recognition have recently been proposed, e.g. DollarStreet \cite{NEURIPS2022_5474d9d4}, GeoNet \cite{kalluri2023geonet}, GeoDE \cite{ramaswamy2023beyond}, GeoYFCC \cite{dubey2021adaptive}, and OpenImages-Extended \cite{chi2019crowdsourcing}.  
        Interestingly, \cite{prabhu2022can, kalluri2023geonet} demonstrate that \emph{traditional methods in unsupervised domain adaptation} \cite{ganin2015unsupervised,long2015learning,long2018conditional,jin2020minimum,xu2019larger,zhang2019bridging,saito2018maximum,wei2021toalign,kalluri2022memsac} which seek to bridge gaps based on visual features alone, \emph{do not effectively address geographical domain shift}. They achieve negligible gains (e.g. 0.14 for \cite{ganin2015unsupervised} in \cite{prabhu2022can}) or often drops in performance (e.g. all methods tested in \cite{kalluri2023geonet}), compared to just using the source model. 
        Attempts to specifically address geographical robustness are limited: 
        \cite{wang2020train} 
        corrects for differences in the sizes of cars, 
        \cite{dubey2021adaptive} proposes 
        a discriminative domain embedding from target data, and
        GiVL \cite{yin2023givl} 
        pretrains with knowledge from Wikipedia. In contrast, our descriptive knowledge regularization works for different categories (not just cars); we do not require target domain data to achieve gains cross-geography; we explore the strong capabilities of LLMs to gather relevant knowledge; and we propose lightweight adaptation through soft prompting (unlike GiVL's expensive pretraining).

    \noindent \textbf{Vision-language (VL) models} 
        \cite{radford2021learning, jia2021scaling, li2021align, li2023blip, yu2022coca} excel on a variety of tasks. CLIP \cite{radford2021learning} shows 
        impressive zero-shot object recognition across different settings. Yet \emph{its performance given geographical shift is less apparent}. GeoNet \cite{kalluri2023geonet} only shows finetuned performance, 
        which is expensive given CLIP's large scale. GeoDE \cite{ramaswamy2023beyond} only shows zero-shot inference with CLIP's default prompts. Neither work evaluates descriptive knowledge or soft prompting. 
                
    \noindent \textbf{Learning soft textual prompts.} 
        Several recent works to adapt CLIP have focused on parameter and data efficiency using linear probing \cite{radford2021learning} and prompting \cite{zhou2022learning,jia2022visual,khattak2023maple}.
        Soft textual prompting (\eg CoOp \cite{zhou2022learning}) is notable as it optimizes class text embeddings (without manual tuning), which we hypothesize is critical to adequately adapt for geographical robustness.
        As CoOp overfits on base (seen) classes, CoCoOp \cite{Zhou_2022_CVPR} proposes to condition prompts on the image for better generalizability. 
        KgCoOp \cite{yao2023visual} alternatively guides learned prompt embeddings towards CLIP's manual prompt embeddings through a distance constraint to avoid degradation on unseen classes. Our approach also uses a distance constraint, but it differs from \cite{yao2023visual} with the purpose of regularizing learned prompt representations for cross-\emph{geography} generalization instead of the base-to-new-\emph{class} setting. We also show novel benefits of regularization when used with an \textit{ensemble} of CLIP's internal geographical knowledge and \textit{external} geographical descriptive knowledge. Our approach notably outperforms each of CoOp, CoCoOp, and KgCoOp by at least +2.8 accuracy on target countries in Africa in DollarStreet.   
        External knowledge aids unseen classes in 
        KAPT \cite{kan2023knowledge}, but \textit{not 
        with respect to geographical knowledge}. 
        Prompt tuning for adaptation has been tested in \cite{shu2022test,ge2022domain}, but not with descriptive knowledge. 
        
      \noindent \textbf{Knowledge probed from large language models}       
      like \cite{brown2020language,team2022chatgpt,openai2023gpt4, choi2023chatgpt} 
      has been used for visual reasoning~\cite{yang2023mmreact}, embodied agent planning~\cite{huang2022language,song2023llmplanner}, 
    and to generate additional context for VLM class prompts in object recognition \cite{menon2022visual,pratt2023does}. We uniquely probe LLMs for 
    \textit{distinguishable visual descriptions for the same object class across different geographical regions}.
    We are also the first to incorporate geographical knowledge from LLMs into soft prompting.

  

\section{Approach}
\label{sec:approach}
    
    We investigate geographical shift in object recognition with VLMs. We posit that the manner in which classes are \textit{described} is critical due to cross-geography shifts in design and context. We also hypothesize that CLIP's default class representations elicited through ``a photo of a/an $<$object$>$'' prompts may not adequately represent classes around the world. Instead, they may be more aligned to high-resource geographies due to Internet-based training data. 
    Optimizing representations (with soft prompts) on a specific geography (\eg Europe) may exacerbate a lack of robustness. 
    Our main idea (Fig.~\ref{approach_fig}) is to incorporate object-related geographical knowledge into prompting to ensure model robustness in different regions.
    We outline our mechanism to obtain geography-specific context by probing CLIP's internal knowledge and an external LLM's descriptive knowledge. We further propose geography knowledge regularization to ensure soft prompts do not overfit when training data is limited to certain geographies.  
    
    \noindent \textbf{Preliminaries.}
    We consider object recognition on a dataset $\mathcal{S}$ containing a class set $\mathcal{C}$ (size $N_c$) over a set of geographies $\mathcal{G}$. 
    We consider a geography $g$ to be either a country or continent. Our VLM is CLIP \cite{radford2021learning}, with an image encoder $f$ and language encoder $h$. We incorporate knowledge of each geography $g$ into prompting using (1) zero-shot inference or (2) soft textual prompting. Prompts are defined as $\boldsymbol{t}$ (each is a set of tokens), and class embeddings $\boldsymbol{w}$ are calculated as $h(\boldsymbol{t})$. We refer to CLIP's default prompt  ``a photo of a/an $<$object$>$'' for a class $c$ as $\boldsymbol{t}_c^{\text{default}}$.

    \subsection{Geographical Knowledge Probing}
    \label{sec:geo_probe}

        \textbf{Probing CLIP's internal geographical knowledge.} 
            Our first strategy of investigation is to augment CLIP's manual prompts to include country names, as we surmise that some of the resulting class representations may be better aligned to how categories present in different regions. \cite{basu2023inspecting} inspires this hypothesis, showing that adding country names to image generation prompts can achieve gains in geographical representativeness. However, it is an open question whether adding country names in prompts improves recognition. We define the setting \textbf{CountryInPrompt}, using the prompt $\boldsymbol{t}_c^{\text{CountryInPrompt}}$ with template ``a photo of a/an $<$object$>$ in $<$country$>$'', \eg 
            ``a photo of a stove in Burundi.'' 
    
        \noindent \textbf{Probing external LLM geographical knowledge.} 
            As CLIP may not have sufficient knowledge of objects in some regions, we consider further augmenting prompts with \textit{external} knowledge.
            Motivated by probing LLMs for general attribute-based object descriptions \cite{menon2022visual, pratt2023does} (\eg a \textit{tiger} with ``stripes and sharp teeth''), we probe GPT-3 (\textit{davinci-003}) for \textit{geography-specific} descriptions of object styles, contexts, and materials.\footnote{We found ChatGPT to perform worse than GPT-3, also found in \cite{pratt2023does}.} We reason that since LLMs are trained on large information sources (\eg CommonCrawl \cite{commoncrawl}, WebText \cite{radford2019language},  Wikipedia \cite{wikipedia}), they may have knowledge about how an object presents in a region due to climate, economics, and/or cultural factors. For instance, \textit{roofs} may sometimes be ``thatched'' in tropical and temperate climates, and \textit{cutlery} may sometimes be made of ``bamboo'' in areas with bamboo forests. Our goal is unique vs. \cite{menon2022visual, pratt2023does} in that we explore descriptive knowledge differences for the same class to address domain shifts across regions. 
            
            \noindent \textit{Acquiring knowledge.} We follow \cite{menon2022visual}, but 
            instead of gathering one set of feature descriptors $\mathcal{D}(c)$ for each $c$, we collect sets \emph{per country}.
            For each class $c$ and geography $g$, we prompt the LLM to generate descriptor lists $\mathcal{D}_g(c)$, using a template consisting of an example question, answer, and format. We use 1-shot prompting to show how to capture geographically representative object designs and contexts. Our prompt exemplifies this below, using the descriptors for Japanese \textit{ofuro}\begin{CJK}{UTF8}{min} (お風呂\end{CJK}, \textit{bathtub}):
    
            \begin{quote}
                \small \textbf{Q:} What are useful features for distinguishing a \underline{bathtub} in a photo that I took in \underline{Japan}?\\
                \textbf{A:} There are several useful visual features to tell there is a \underline{bathtub} in a photo that I took in \underline{Japan}:
                \\ - short in length and deep 
                \\ - square shape 
                \\ - wooden, plastic, or steel material
                \\ - white or brown color 
                \\ - benches on side 
                \\ - next to shower  \\
                \textbf{Q:} What are useful features for distinguishing $<$category$>$ in a photo that I took in $<$country$>$?\\
                \textbf{A:} There are several useful visual features to tell there is/are $<$category$>$ in a photo that I took in $<$country$>$:
            \end{quote}

            \noindent \textit{Using knowledge.}
            To convert LLM outputs to CLIP prompts, each descriptor $d$ in $\mathcal{D}_g(c)$ serves in a prompt $\boldsymbol{t}_{c,d}$.  The format of $\boldsymbol{t}_{c,d}$ is ``a photo of a/an $<$object$>$ which (is/has/etc.) $<$descriptor$>$". The setting where geography-specific LLM descriptors are used in prompting is referred to as \textbf{CountryLLM} (prompts $\boldsymbol{t}_{c,d}^{\text{CountryLLM}}$), while \cite{menon2022visual} is \textbf{GeneralLLM} (prompts $\boldsymbol{t}_{c,d}^{\text{GeneralLLM}}$). To perform zero-shot inference on an image $I$, each class score $s(c, I)$ is computed using the average of CLIP logits $\phi(I,d)$ over each $d$ in the set $\mathcal{D}$. For GeneralLLM, the score is calculated as:
            \begin{equation}
                \label{score_orig}
                s(c, I) = \frac{1}{|\mathcal{D}(\textit{c})|}\sum_{d \in \mathcal{D}(\textit{c})} \phi(I,d)
            \end{equation}

            \noindent For CountryLLM, we use the geo-specific set: 
            \begin{equation}
                \label{score_country}
                s(c, I, g) = \frac{1}{|\mathcal{D}_g(\textit{c})|}\sum_{d \in \mathcal{D}_g(\textit{c})} \phi(I,d)
            \end{equation}

        \noindent The argmax of $s$ with respect to $c$ is taken as the prediction. Due to averaging over descriptor scores, 
        not every descriptor needs to strongly activate in a correct prediction. The model therefore can account for diverse features of objects within a geography. These descriptors effectively serve as \textit{complements} 
        to CLIP's default knowledge of class names.
        
        \noindent \textbf{Combining knowledge.} 
            Our third method of exploration, 
            \textbf{CountryInPrompt+LLM}, combines both CLIP's internal knowledge and LLM external knowledge. The prompt template $\boldsymbol{t}_{c,d}^{\text{CountryInPrompt+LLM}}$ is ``a photo of a/an $<$object$>$ in $<$country$>$ which (is/has/etc.) $<$descriptor$>$". 
        





    \subsection{Regularizing Soft Prompts via Geo Knowledge}
    \label{sec:geo_reg}

        \textbf{Adaptation scenario.} In practice, one may want to further optimize a VLM for a downstream task. To update a model effectively, one promising strategy is soft textual prompting. It is parameter-efficient \cite{zhou2022learning} and avoids feature distortion unlike finetuning \cite{kumar2022fine}. Its mechanism is to learn \emph{context} parameters that directly change the \emph{class text embeddings} used in inference. We posit that learning context on a dataset with limited diversity (\eg just Europe) may tailor these class representations to the region and overfit. To investigate cross-geography generalization when using soft prompting, we pose a domain generalization scenario where we aim to learn only from a high-resource \textit{source set} of countries and generalize to a \textit{target set} of countries at inference time. A method that performs well in this setting could provide a viable alternative to few-shot target training when acquiring target data for training is not feasible. 
    
        \noindent \textbf{Soft prompts.} 
        Our idea is to learn soft prompts while constraining the class text embeddings to be close to geographical knowledge of objects \textit{outside} of source geographies. In this way, we hope to learn class representations that are more applicable to the rest of the world. Building from CoOp \cite{zhou2022learning}, we assume there is a text prompt $\boldsymbol{t}_c$ for each class $c$. All prompts share $M$ context vectors (each denoted [\text{V}]$_m$), which are the same size as the word embeddings (i.e. 512-D) and precede a class name token [CLASS$_c$]: 
            \begin{equation} 
                \label{soft_prompt_basis}
                {\boldsymbol{t}_c = [\text{V}]_1 [\text{V}]_2 ... [\text{V}]_M [\text{CLASS$_c$}]}
            \end{equation}

            \noindent The respective class text embedding $\boldsymbol{w}_{c}$ is produced as $h(\boldsymbol{t}_c)$, forwarding the prompt through the text encoder. 
            Learning proceeds by minimization of cross-entropy, for image $k$ with features $\boldsymbol{f}_k$, using ground-truth \emph{source} labels $y_{k,c}$ and temperature $\tau$:
            \begin{equation} 
                \label{ce}
                {\mathcal{L}_{ce} = - \sum_{c=1}^{N_c} y_{k,c} \log \frac{\exp (\cos{(\boldsymbol{w}_{c}, \boldsymbol{f}_k)}/ \tau)}{\sum_{j=1}^{N_c} \exp (\cos{(\boldsymbol{w}_{j}, \boldsymbol{f}_k)}/ \tau)
                }}
            \end{equation}

            \noindent \textbf{Geography knowledge regularization (gkr).} We minimize the cosine distance of normalized class embedding $\boldsymbol{w_c}$ 
            and overall target class knowledge $\boldsymbol{k}_{c}^{tgt}$, over all $c$:
            \begin{equation} 
                \label{loss_reg}
                {\mathcal{L}_{gkr} = 1 - \frac{1}{N_c} \sum_{c=1}^{N_c} \cos(\boldsymbol{w_c}, \boldsymbol{k}_c^{tgt}) }
            \end{equation}

            \noindent \textbf{Geo knowledge ensemble.} 
            To define $\boldsymbol{k}_{c}^{tgt}$, we identify that a model may be deployed in various locations. Therefore, we define a \textit{target} geography set $\mathcal{G}_t$, which can practically be thought of as the countries that a model may be deployed in that are not in the training set $\mathcal{D}$ (\eg Africa, Asia, Americas in $\mathcal{G}_t$ if only Europe in $\mathcal{D}$). 
            Then for each geography $g$ in $\mathcal{G}_t$, we define the corresponding class knowledge $\boldsymbol{k}_{c}^{g}$ as:
            \begin{equation} 
                \label{knowledge_per_class}
                {\boldsymbol{k}_{c}^{g} = \frac{1}{|\mathcal{D}_g(c)|} \sum_{d\in \mathcal{D}_g(c)}} \boldsymbol{w}_{c,d}^{\text{CountryInPrompt+LLM}}
            \end{equation}
            This is defined analogously for CountryInPrompt and CountryLLM.
            The final regularization target $\boldsymbol{k}_{c}^{tgt}$ 
            for class $c$ 
            aggregates the set's geographical knowledge:
            \begin{equation} 
                \label{overall_knowledge}
                {\boldsymbol{k}_{c}^{tgt} = \frac{1}{|\mathcal{G}_t|} \sum_{g \in \mathcal{G}_t} \boldsymbol{k}_{c}^{g}}
            \end{equation}

\noindent While the loss formulation includes cosine distance like KgCoOp \cite{yao2023visual}, it serves a different purpose: we regularize for \textit{cross-geography domain generalization}, while KgCoOp regularizes for base-class-to-new-class inference. Our method outperforms KgCoOp in cross-geography generalization due to its use of geo-specific knowledge. 

           \noindent \textbf{Overall loss.} 
            The final loss $\mathcal{L}$ for learning soft prompts, where $\lambda$ controls the strength of regularization, is:
            \begin{equation} 
                \label{total_loss}
                {\mathcal{L} = \mathcal{L}_{ce} + \lambda \mathcal{L}_{gkr} }
            \end{equation}

             \setlength{\tabcolsep}{1pt}

\begin{table*}[t]

    \begin{center}
    \renewcommand{\arraystretch}{0.82}

    \begin{tabular}{c|c||cc|cc|cc|cc|cc||cc|cc|cc|cc|cc}
        
        \hline 
        & & \multicolumn{10}{c||}{\textbf{\small \textit{Top-1 Accuracy}}} 
        & \multicolumn{10}{c}{\textbf{\small \textit{Top-3 Accuracy}}} 
        \\
        
        \textbf{\small Encoder} & \textbf{\small Prompting Method} & \multicolumn{2}{c}
        {\textbf{\small Europe}} & \multicolumn{2}{c}{\textbf{\small Africa}} &\multicolumn{2}{c}{\textbf{\small Asia}} & \multicolumn{2}{c}{\textbf{\small Americas}} & \multicolumn{2}{c||}{\textbf{\small Total}} & \multicolumn{2}{c}
        {\textbf{\small Europe}} & \multicolumn{2}{c}{\textbf{\small Africa}} &\multicolumn{2}{c}{\textbf{\small Asia}} & \multicolumn{2}{c}{\textbf{\small Americas}} & \multicolumn{2}{c}{\textbf{\small Total}} \\
         & 
         &  \multicolumn{1}{c}{\small Acc} 
         & \multicolumn{1}{c}{\small $\Delta$} 
         &  \multicolumn{1}{c}{\small Acc} 
         & \multicolumn{1}{c}{\small $\Delta$} 
         & \multicolumn{1}{c}{\small Acc} 
         & \multicolumn{1}{c}{\small $\Delta$} 
         &  \multicolumn{1}{c}{\small Acc} 
         & \multicolumn{1}{c}{\small $\Delta$} 
         & \multicolumn{1}{c}{\small Acc} 
         & \multicolumn{1}{c||}{\small $\Delta$} 
             &  \multicolumn{1}{c}{\small Acc} 
         & \multicolumn{1}{c}{\small $\Delta$} 
         &  \multicolumn{1}{c}{\small Acc} 
         & \multicolumn{1}{c}{\small $\Delta$} 
         & \multicolumn{1}{c}{\small Acc} 
         & \multicolumn{1}{c}{\small $\Delta$} 
         &  \multicolumn{1}{c}{\small Acc} 
         & \multicolumn{1}{c}{\small $\Delta$} 
         & \multicolumn{1}{c}{\small Acc} 
         & \multicolumn{1}{c}{\small $\Delta$} 
         \\
         
    \hline\hline \rowcolor{LLGray}
        \cellcolor{white} \small ViT-B/32  
        &  \small Zero-Shot CLIP \cite{radford2021learning} 
        & \small 59.1 & \scriptsize - 
        &  \small43.7 & \scriptsize - 
        &  \small50.8 & \scriptsize - 
        & \small \textbf{55.3} & \scriptsize - 
        & \small 51.7 & \scriptsize -

        & \small 81.1 & \scriptsize - 
        & \small 64.8 & \scriptsize - 
        & \small 72.3 & \scriptsize - 
        & \small \textbf{77.4} & \scriptsize - 
        & \small 73.7 & \scriptsize -
        
        \\
        
         \rowcolor{LLGray}
        \small \cellcolor{white} & \small GeneralLLM \cite{menon2022visual}
         & \small 57.3 & \scriptsize \color{red}{-1.8} 
         & \small 44.3 & \scriptsize \color{Greenish}{+0.6} 
         & \small 50.9 & \scriptsize \color{Greenish}{+0.1} 
         & \small 54.6 & \scriptsize \color{red}{-0.7} 

         & \small 51.4 & \scriptsize \color{red}{-0.3}
        & \small 78.8 & \scriptsize \color{red}{-2.3} 
        & \small 64.5 & \scriptsize \color{red}{-0.3} 
        & \small 72.1 & \scriptsize \color{red}{-0.2} 
        & \small 75.7 & \scriptsize \color{red}{-1.7}
        & \small 73.0 & \scriptsize \color{red}{-0.7}
         \\

         &  \small CountryInPrompt 

         & \small 57.5 & \scriptsize \color{red}{-1.6} 
         & \small 45.2 & \scriptsize \color{Greenish}{+1.5} 
         & \small 51.9 & \scriptsize \color{Greenish}{+1.1} 
         & \small 55.0 & \scriptsize \color{red}{-0.3} 

         & \small 52.1 & \scriptsize \color{Greenish}{+0.4}

        & \small 80.2 & \scriptsize \color{red}{-0.9} 
        & \small 65.5 & \scriptsize \color{Greenish}{+0.7}
        & \small 73.3 & \scriptsize \color{Greenish}{+1.0}
        & \small 76.9 & \scriptsize \color{red}{-0.5}
        & \small 73.9 & \scriptsize \color{Greenish}{+0.2}
         \\
         
         &  \small CountryLLM 
         & \small 59.4 & \scriptsize \color{Greenish}{+0.3} 
         & \small 45.2 & \scriptsize \color{Greenish}{+1.5} 
         & \small 52.1 & \scriptsize \color{Greenish}{+1.3} 
         & \small \textbf{55.3} & \scriptsize \color{black}{0.0} 

         & \small 52.6 & \scriptsize \color{Greenish}{+0.9}
        & \small 80.9 & \scriptsize \color{red}{-0.2}
        & \small 66.4 & \scriptsize \color{Greenish}{+1.6}
        & \small \textbf{73.6} & \scriptsize \color{Greenish}{+1.3}
        & \small \textbf{77.4} & \scriptsize \color{black}{0.0}
        & \small 74.6 & \scriptsize \color{Greenish}{+0.9}
        \\

         & \small CountryInPrompt+LLM  
          &  \small\textbf{60.8} & \scriptsize \color{Greenish}{+1.7}
         & \small \textbf{45.3} & \scriptsize \color{Greenish}{+1.6} 
         & \small \textbf{52.2} & \scriptsize \color{Greenish}{+1.4}
         & \small 55.0 & \scriptsize \color{red}{-0.3} 
         & \small \textbf{52.8} & \scriptsize \color{Greenish}{+1.1}

        & \small \textbf{81.5} & \scriptsize \color{Greenish}{+0.4} 
        & \small \textbf{67.4} & \scriptsize \color{Greenish}{+2.6}
        & \small \textbf{73.6} & \scriptsize \color{Greenish}{+1.3}
        & \small 76.7 & \scriptsize \color{red}{-0.7} 
        & \small \textbf{74.7} & \scriptsize \color{Greenish}{+1.0}
        \\
         
    \hline\hline \rowcolor{LLGray}
         \cellcolor{white} \small ViT-B/16  &  \small Zero-Shot CLIP \cite{radford2021learning}  
        & \small 64.3 & \small - 
       &  \small 46.9 & \small - 
       & \small  53.9 & \small - 
       &  \small \textbf{60.1} & \small - 
       &  \small 55.5 & \small - 

        & \small 84.3 & \small - 
        & \small 69.3 & \small - 
        & \small 75.9 & \small - 
        & \small 81.1 & \small - 
        & \small 77.2 & \small -
       \\
       
         \rowcolor{LLGray}
        \cellcolor{white} & \small GeneralLLM \cite{menon2022visual} 
          & \small  64.2 & \scriptsize \color{red}{-0.1} 
         & \small 48.8 & \scriptsize \color{Greenish}{+1.9} 
         & \small \textbf{56.0} & \scriptsize \color{Greenish}{+2.1} 
         & \small 58.5 & \scriptsize \color{red}{-1.6} 

         & \small 56.8 & \scriptsize \color{Greenish}{+1.3}
        & \small 83.9 & \scriptsize \color{red}{-0.4} 
        & \small 71.1 & \scriptsize \color{Greenish}{+1.8} 
        & \small 76.3 & \scriptsize \color{Greenish}{+0.4}
        & \small 80.4 & \scriptsize \color{red}{-0.7}
        & \small 77.9 & \scriptsize \color{Greenish}{+0.7}
        \\

         & \small CountryInPrompt 
         & \small 63.9 & \scriptsize \color{red}{-0.4} 
         & \small 49.6 & \scriptsize \color{Greenish}{+2.7} 
         & \small 55.7 & \scriptsize \color{Greenish}{+1.8} 
         & \small 59.3 & \scriptsize \color{red}{-0.8} 

         &\small 56.6 & \scriptsize \color{Greenish}{+1.1}
        & \small 84.0 & \scriptsize \color{red}{-0.3}
        & \small 71.3 & \scriptsize \color{Greenish}{+2.0}
        & \small 76.5 & \scriptsize \color{Greenish}{+0.6}
        & \small 80.0 & \scriptsize \color{red}{-1.1}
        & \small 77.7 & \scriptsize \color{Greenish}{+0.5}
         \\

         & \small  CountryLLM  
         &  \small65.2 & \scriptsize \color{Greenish}{+0.9} 
         & \small 49.6 & \scriptsize \color{Greenish}{+2.7} 
         & \small55.6 & \scriptsize \color{Greenish}{+1.7} 
         & \small59.7 & \scriptsize \color{red}{-0.4} 

         & \small 57.0 & \scriptsize \color{Greenish}{+1.5}

        & \small 84.3 & \scriptsize \color{black}{0.0}
        & \small 71.8 & \scriptsize \color{Greenish}{+2.5}
        & \small \textbf{77.5} & \scriptsize \color{Greenish}{+1.6}
        & \small \textbf{81.5} & \scriptsize \color{Greenish}{+0.4}
        & \small \textbf{78.8} & \scriptsize \color{Greenish}{+1.6}
         \\
         
         &  \small CountryInPrompt+LLM 
         & \small \textbf{65.5} & \scriptsize \color{Greenish}{+1.2} 
         & \small\textbf{50.8} & \scriptsize \color{Greenish}{+3.9} 
         & \small \textbf{56.0} & \scriptsize \color{Greenish}{+2.1} 
         &  \small 59.7 & \scriptsize \color{red}{-0.4} 
         & \small \textbf{57.4} & \scriptsize \color{Greenish}{+1.9}

         & \small \textbf{85.5} & \scriptsize \color{Greenish}{+1.2}
        & \small \textbf{72.5} & \scriptsize \color{Greenish}{+3.2}
        & \small 77.0 & \scriptsize \color{Greenish}{+1.1}
        & \small 80.9 & \scriptsize \color{red}{-0.2}
        & \small 78.7 & \scriptsize \color{Greenish}{+1.5}
         \\
    \hline
    \hline
    \rowcolor{LLGray}
         \cellcolor{white} \small RN50 
        & \small Zero-Shot CLIP \cite{radford2021learning} 
        & \small 53.0 & \scriptsize - 
        & \small 38.0 & \scriptsize - 
        & \small 44.4 & \scriptsize - 
        & \small 49.8 & \scriptsize - 

        & \small 45.7 & \scriptsize -
        & \small 76.5 & \scriptsize - 
        & \small 60.2 & \scriptsize - 
        & \small 66.4 & \scriptsize - 
        & \small \textbf{72.7} & \scriptsize - 
        & \small 68.1 & \scriptsize -
       \\

         \rowcolor{LLGray}
        \small \cellcolor{white} & \small GeneralLLM \cite{menon2022visual}
         & \small 55.5 & \scriptsize \color{Greenish}{+2.5} 
         & \small40.9 & \scriptsize \color{Greenish}{+2.9} 
         &\small 46.9 & \scriptsize \color{Greenish}{+2.5} 
         &  \small50.3 & \scriptsize \color{Greenish}{+0.5} 

         & \small 47.9 & \scriptsize \color{Greenish}{+2.2}
        & \small 76.0 & \scriptsize \color{red}{-0.5}
        & \small 61.2 & \scriptsize \color{Greenish}{+1.0} 
        & \small 67.7 & \scriptsize \color{Greenish}{+1.3}
        & \small 71.1 & \scriptsize \color{red}{-1.6}
        & \small 68.6 & \scriptsize \color{Greenish}{+0.5}
         \\

         & \small CountryInPrompt 

         & \small 54.5 & \scriptsize \color{Greenish}{+1.5} 
         & \small \textbf{43.4} & \scriptsize \color{Greenish}{+5.4} 
         & \small 47.0 & \scriptsize \color{Greenish}{+2.6} 
         & \small 50.8 & \scriptsize \color{Greenish}{+1.0} 

         & \small 48.4 & \scriptsize \color{Greenish}{+2.7}

        & \small 76.0 & \scriptsize \color{red}{-0.5}
        & \small \textbf{64.0} & \scriptsize \color{Greenish}{+3.8} 
        & \small 68.7 & \scriptsize \color{Greenish}{+2.3}
        & \small \textbf{72.7} & \scriptsize \color{black}{0.0}
        & \small \textbf{70.0} & \scriptsize \color{Greenish}{+1.9}
        \\

         &  \small CountryLLM 
         & \small 56.2 & \scriptsize \color{Greenish}{+3.2}
         &  \small41.1 & \scriptsize \color{Greenish}{+3.1} 
         & \small  47.3 & \scriptsize \color{Greenish}{+2.9} 
         & \small50.4 & \scriptsize \color{Greenish}{+0.6} 

         & \small 48.3 & \scriptsize \color{Greenish}{+2.6}

        & \small \textbf{77.2} & \scriptsize \color{Greenish}{+0.7}
        & \small 62.5 & \scriptsize \color{Greenish}{+2.3}
        & \small \textbf{68.8} & \scriptsize \color{Greenish}{+2.4}
        & \small 72.4 & \scriptsize \color{red}{-0.3}
        & \small \textbf{70.0} & \scriptsize \color{Greenish}{+1.9}
        \\
         
         &  \small CountryInPrompt+LLM 
         & \small \textbf{56.4} & \scriptsize \color{Greenish}{+3.4}
         & \small 43.0 & \scriptsize \color{Greenish}{+5.0} 
         & \small \textbf{48.0} & \scriptsize \color{Greenish}{+3.6} 
         & \small \textbf{50.9} & \scriptsize \color{Greenish}{+1.1} 

         &\small\textbf{49.1} & \scriptsize \color{Greenish}{+3.4}

        & \small 76.7 & \scriptsize \color{Greenish}{+0.2}
        & \small 63.1 & \scriptsize \color{Greenish}{+2.9}
        & \small 68.3 & \scriptsize \color{Greenish}{+1.9}
        & \small 71.1 & \scriptsize \color{red}{-1.6} 
        & \small 69.4 & \scriptsize \color{Greenish}{+1.3}
         \\ 
    \hline
    \end{tabular}
    \end{center}
    \vspace{-5mm}
    \caption{\textbf{Zero-shot CLIP inference with descriptive knowledge prompts, top-1/3 balanced accuracy (Acc) on DollarStreet.} Strategies to capture CLIP's internal country knowledge (CountryInPrompt), external LLM country knowledge (CountryLLM), and their combination (CountryInPrompt+LLM), often improve vs. the zero-shot CLIP baseline (prompt ``a photo of a/an $<$object$>$''), especially on Africa and Asia; gains in \color{Greenish}{green}\color{black}, drops in \color{red}{red}\color{black}. CountryLLM notably outperforms the GeneralLLM \cite{menon2022visual} baseline. }
    \label{table:zero_shot}
\end{table*}

     \setlength{\tabcolsep}{4pt}

\begin{table*}[t]
    \begin{center}
        \renewcommand{\arraystretch}{0.82}

    \begin{tabular}{c|c|cc|cc|cc|cc|cc}
        \hline
         && \multicolumn{2}{c|}{\textit{\small \textbf{Source}}} 
         & \multicolumn{8}{c}{\textit{\small \textbf{Target}}}  \\
        
         \textbf{Encoder} &
         \textbf{Prompting Method} 
         & \multicolumn{2}{c|}{\textbf{Europe}} 
         & \multicolumn{2}{c}{\textbf{Africa}} 
         & \multicolumn{2}{c}{\textbf{Asia}} 
         & \multicolumn{2}{c}{\textbf{Americas}} 
         & \multicolumn{2}{c}{\textbf{Total}} \\
        
         & & \multicolumn{1}{c}{Acc} 
         & \multicolumn{1}{c|}{\small $\Delta$} 
         & \multicolumn{1}{c}{Acc} 
         & \multicolumn{1}{c}{\small $\Delta$} 
         & \multicolumn{1}{c}{Acc} 
         & \multicolumn{1}{c}{\small $\Delta$} 
         & \multicolumn{1}{c}{Acc} 
         & \multicolumn{1}{c}{\small $\Delta$}  
         & \multicolumn{1}{c}{Acc}
         & \multicolumn{1}{c}{\small $\Delta$} 
        \\
    
    \hline\hline
    
        \rowcolor{LLGray}\cellcolor{white}  ViT-B/16 &
        
          CoOp \cite{zhou2022learning} 
         &  \color{gray}{72.2} & \small - 
         & 53.9 & \small - 
         &  61.5 & \small - 
         &  68.6 & \small - 
         &  61.7 & \small - 
         \\
         
         \rowcolor{LLGray} \cellcolor{white} & CoCoOp \cite{Zhou_2022_CVPR} 
         &  \color{gray}{73.2} & \small - 
          &  54.3 & \small - 
         &  61.2 & \small - 
         & 68.3 & \small - 
         &  61.4 & \small - 
         \\
         
        
         \rowcolor{LLGray} \cellcolor{white} &  KgCoOp  \cite{yao2023visual} 
         &  \color{gray}{73.1} & \small - 
         &  54.4 & \small - 
         & 62.6 & \small - 
         &  68.7 & \small - 
         &  62.4 & \small - 
         \\
         
         & CountryInPrompt Reg
         &  \color{gray}{71.8} & \textit{\small \color{gray}{-1.4}} 
         &  56.8 & \small \color{Greenish}{+2.4} 
         &  63.0 & \small \color{Greenish}{+0.4} 
         &  69.8 & \small \color{Greenish}{+1.1} 
         &  63.5 & \small \color{Greenish}{+1.1} 
         \\
         
         & CountryLLM Reg
         &  \color{gray}{73.2} & \textit{\small \color{gray}{0.0}} 
         &  55.6 & \small \color{Greenish}{+1.2} 
         & 63.0 & \small \color{Greenish}{+0.4}
         &  70.0 & \small \color{Greenish}{+1.3} 
         &  63.2 & \small \color{Greenish}{+0.8} 
         \\
         
         & CountryInPrompt+LLM Reg
         &  \color{gray}{\textbf{73.6}} & \small \textit{\color{gray}{+0.4}} 
         & \textbf{57.2} & \small \color{Greenish}{+2.8} 
         & \textbf{63.8} & \small \color{Greenish}{+1.2} 
         & \textbf{70.3} & \small \color{Greenish}{+1.6} 
         & \textbf{64.0} & \small \color{Greenish}{+1.6} 
         \\
         
    \hline\hline
        
        \rowcolor{LLGray} \cellcolor{white}  RN50 
        
        &  CoOp \cite{zhou2022learning} 
        & \color{gray}{64.6} & \small - 
        &  45.2 & \small - 
        &51.6 & \small - 
        &  59.5 & \small - 
        &  52.2 & \small - 
        \\
        
        \rowcolor{LLGray} \cellcolor{white} &  CoCoOp \cite{Zhou_2022_CVPR} 
        & \color{gray}{62.9} & \small - 
        & 44.5 & \small - 
        & 51.0 & \small - 
        & 58.3 & \small - 
        &  51.4 & \small - 
        \\
        
        
        \rowcolor{LLGray} \cellcolor{white}  & KgCoOp \cite{yao2023visual} 
        & \color{gray}{63.5} & \small - 
        & 46.3 &\small - 
        &  53.9 & \small - 
        &  \textbf{60.5} & \small - 
        &  53.9 & \small - 
        
        \\
        
        &  CountryInPrompt Reg 
        &  \color{gray}{63.5} & \textit{\small \color{gray}{-1.1}}
        &  48.0 & \small \color{Greenish}{+1.7} 
        &  53.9 & \small \color{black}{0.0} 
        &  60.3 & \small \color{red}{-0.2} 
        & 54.3 & \small \color{Greenish}{+0.4} 
        
        \\
        
         &  CountryLLM Reg 
         &  \color{gray}{64.5} & \small \textit{\color{gray}{-0.1}} 
         & 47.4 & \small \color{Greenish}{+1.1} 
         &  54.2 & \small \color{Greenish}{+0.3}
         &  59.9 & \small \color{red}{-0.6} 
         & 54.3 & \small \color{Greenish}{+0.4} 
         
         \\
         
         & CountryInPrompt+LLM Reg   
         & \color{gray}{\textbf{65.5}} & \textit{\small \color{gray}{+0.9}} 
         &  \textbf{48.1} & \small \color{Greenish}{+1.8} 
         & \textbf{54.5} & \small \color{Greenish}{+0.6} 
         &  60.4 & \small \color{red}{-0.1} 
        
        & \textbf{54.8} & \small \color{Greenish}{+0.9} 
         
         \\

    \hline
    \end{tabular}
    \end{center}
    \vspace{-5mm}
    \caption{\textbf{Regularizing soft prompts with geographical knowledge, top-1 bal. acc. on DollarStreet.} We emphasize that our regularization aims to improve \textbf{target} performance, rather than source (\color{gray}{gray}, \textit{italicized}\color{black}). 
    \color{Greenish}{Gains}\color{black}/\color{red}{drops }\color{black} are shown vs. the \textit{best} of soft prompt baselines (shaded). CountryInPrompt+LLM Reg achieves notable gains in target, especially on Africa. Methods use 16 shots per class.}
    \label{table:soft_prompt}
\end{table*}
\section{Experimental Setup}
    \label{sec:exp}

        
    \noindent \textbf{Datasets.} 
    We use \textit{DollarStreet} \cite{NEURIPS2022_5474d9d4}, which
    has 38,479 images of household objects across regions (Africa, South/Central/North America, Asia, Europe) and incomes. The classes may represent abstract concepts (\eg \textit{most loved toys}), so we narrow focus to 95 object classes. We merge especially close categories (\textit{light sources by bed/in living room}) and ignore multi-label examples, resulting in 23,114 total images. For zero-shot inference, the entire set is used. For training, the source is Europe, and the target is Americas, Asia, and Africa. 20\% of source data, stratified based on class proportions, is heldout for testing; target evaluation is on all data from target continents. To set up $\boldsymbol{k}_c^{tgt}$, the 49 target countries in DollarStreet make up $\mathcal{G}_t$. We also use the GeoImNet benchmark of \textit{GeoNet} \cite{kalluri2023geonet}, comprised of 171,692 images across 600 objects from the USA (source) and 78,358 images across the same number in Asia (target). We use existing train-test splits for soft prompt training. For GeoNet, given the relatively large number of categories and inference costs of \textit{davinci-003}, $\boldsymbol{k}_c^{tgt}$ and $\mathcal{G}_t$ use the top 10 most frequent countries in the GeoNet set.

    
    \noindent \textbf{Baselines.} We evaluate geography knowledge regularization vs. CoOp \cite{zhou2022learning}, CoCoOp \cite{Zhou_2022_CVPR} and KgCoOp \cite{yao2023visual}. For zero-shot inference, we evaluate CLIP with default prompts and the classification via description method of \cite{menon2022visual}. 
    
    \noindent \textbf{Metrics.} We report balanced accuracy, which is the average of per-class recall scores. We use this metric to account for class imbalance in both DollarStreet and GeoNet. For zero-shot inference, we also show top-3 accuracy as some similar categories exist (\eg \textit{cooking utensils}, \textit{cutlery}). 
    
    \noindent \textbf{Experimental details.} For all soft prompting experiments, models are trained with 16 shots, context length $M=4$, and for 100 epochs, unless otherwise stated. The class token position follows the soft prompts, and class-shared context is used. Our method uses a batch size of 128 (same as KgCoOp), while the batch sizes for CoOp and CoCoOp follow 
    \cite{yao2023visual} (i.e. 32, and 1 for CoCoOp due to memory limitations). The encoders used for training include ViT-B/16 \cite{dosovitskiy2020image} and ResNet-50 (RN50) \cite{he2016deep} as reported in \cite{yao2023visual}. Both our method and KgCoOp use a regularization weight $\lambda$. We set $\lambda=4$ for DollarStreet, and compare to KgCoOp at $\lambda=4$ (which performs better than KgCoOp's default $\lambda=8$). For GeoNet, we use $\lambda=8$. Training is performed on 1 NVIDIA Quadro RTX A5000 GPU with 24 GB of memory. All reported soft prompt results are averages over 3 runs. For experiments in the zero-shot setting, results are shown over ViT-B/16, ViT-B/32, and RN50 encoders. LLM descriptors for all experiments are generated from the \textit{davinci-003} version of GPT-3, with max tokens 100 and temperature 0.7.

\section{Results}
\label{sec:results}
            
\subsection{Zero-shot CLIP Inference with Geo Knowledge}

We gauge the effectiveness of three zero-shot strategies: 
(1) CountryInPrompt (including countries in prompts to probe CLIP's knowledge), (2) CountryLLM (gathering descriptive knowledge of objects with \textit{davinci-003}), and (3) CountryInPrompt+LLM (using country names and LLM knowledge). We compare to \cite{menon2022visual} (GeneralLLM) and CLIP with manual prompts (i.e. ``a photo of a/an $<$object$>$''). Results on DollarStreet are shown in Table \ref{table:zero_shot}.

\noindent \textbf{Including country names in prompts can improve object recognition, especially in Africa and Asia}. This observation is supported by gains for CountryInPrompt vs. Zero-Shot CLIP, especially in Africa and Asia (up to +5.4 and +2.6 top-1 gains for RN50, resp.). Such differences may occur as country-specific context can align representations closer to these regions, while default prompts do not adequately capture objects around the world (esp. from non-Western regions). In Americas/Europe, adding country names leads to gains with RN50, but slight drops with ViT-B/16 and ViT-B/32. We reason that CLIP's default prompts may be already well-aligned to countries in these regions for those encoders due to overrepresentation in training.


\noindent \textbf{Prompting with country-specific descriptive knowledge from LLMs outperforms general object knowledge}. We observe this from CountryLLM's larger gains over default CLIP than GeneralLLM's for almost all encoders, regions, and metrics. The largest top-1 difference is with ViT-B/32 (in Total, 52.6\% for CountryLLM vs. 51.4\% for GeneralLLM). In top-3 accuracy, the differences for CountryLLM/GeneralLLM in Total are 74.6/73.0 for ViT-B/32, 78.8/77.9 for ViT-B/16, 70.0/68.6 for RN50. These suggest that default non-country-specific knowledge is less adequate for various countries. 
The gains of CountryLLM vs. Zero-Shot CLIP are generally largest on Africa and Asia, as countries in these regions may have greater 
shifts vs. the default prompts, but CountryLLM also performs well on Europe. LLM description in general is less effective in Americas, though Americas has a large proportion of USA images, for which default CLIP may be well-aligned. 
        
\noindent \textbf{There are complementary effects when using CLIP's internal and external LLM geo knowledge.}
This observation is supported by CountryInPrompt+LLM, the combination of CountryInPrompt and CountryLLM, achieving the best Total top-1 performance for every encoder. The Total gains vs. default CLIP are as large as +3.4 (RN50). While CLIP has internal knowledge of country-specific categories, it may be incomplete and imprecise due to limited representation in the image-text training corpus. Adding LLM knowledge, trained on a purely textual corpus, may address some gaps. CountryInPrompt+LLM is notably the top setting in 3/4 regions for each encoder in top-1 accuracy. 

\subsection{Soft Prompting}

We next evaluate geography knowledge regularization (Sec.~\ref{sec:geo_reg}), our method \textit{to improve target performance} by ensuring that soft prompts do not overfit class text representations to a source dataset with limited geographical representativeness (\eg only data from Europe). 
We compare regularization with ensembles of CountryInPrompt, CountryLLM, and CountryInPrompt+LLM prompts vs. state-of-the-art soft prompting methods in Tables \ref{table:soft_prompt}/\ref{table:soft_prompt_2}. 
        
\noindent \textbf{Regularizing soft prompts with target geographical knowledge reduces overfitting to source geographies.} Our method effectively improves the ability of CLIP, \textit{with prompts trained only on images from Europe}, to generalize to target countries. This observation is supported by Total Target gains for CountryInPrompt, CountryLLM, and CountryInPrompt+LLM Reg on DollarStreet (+1.1/0.8/1.6 over the \emph{best soft prompt baseline} for ViT-B/16). Improvements are notable in Africa: CountryInPrompt+LLM achieves +2.8 for ViT-B/16 and +1.8 for RN50. The effectiveness extends to GeoNet in Table \ref{table:soft_prompt_2}: target gains are +1.3 for ViT-B/16 and +1.4 for RN50. The combined strategy works best on target, showing the value of incorporating descriptive knowledge. Since regularization prevents overfitting and potentially optimal source performance, we naturally observe 
\emph{source} drops for CountryInPrompt and CountryLLM in Tables \ref{table:soft_prompt}/\ref{table:soft_prompt_2}. However, CountryInPrompt+LLM in Table \ref{table:soft_prompt} even offers source gains. It is also notable that there are small drops in Americas for RN50, but upon inspection, countries in North America overall have a -0.9 drop, while ones in Central/South America have a +0.8 gain. These results concur with our hypothesis that CLIP is already aligned to countries like the USA. More is in supp., along with experiments varying the source and ensemble.


 \setlength{\tabcolsep}{4pt}

\begin{table}[t]
    \begin{center}
        \renewcommand{\arraystretch}{0.82}

    \begin{tabular}{c|c|cc|cc}
        \hline
         && \multicolumn{2}{c|}{\textit{\small \textbf{Source}}} 
         & \multicolumn{2}{c}{\textit{\small \textbf{Target}}}  \\
        
         \textbf{Encoder} & \textbf{Method} & \multicolumn{2}{c|}{\textbf{USA}} &\multicolumn{2}{c}{\textbf{Asia}}  \\
        
         & & \multicolumn{1}{c}{\small Acc} & \multicolumn{1}{c|}{\small $\Delta$} & 
         \multicolumn{1}{c}{\small Acc} & \multicolumn{1}{c}{\small $\Delta$}    
         
         \\
    
    \hline\hline
    
         \rowcolor{LLGray} \cellcolor{white} ViT-B/16
        
         & \small CoOp \cite{zhou2022learning} 
         & \small \color{gray}{\textbf{58.7}} & \scriptsize -  
         & \small 51.2 & \scriptsize - 
         \\

         \rowcolor{LLGray} \cellcolor{white}  & \small CoCoOp \cite{Zhou_2022_CVPR} 
         & \small \color{gray}{57.7} & \scriptsize - 
         & \small 52.6 & \scriptsize - 
 \\

         \rowcolor{LLGray} \cellcolor{white} & \small KgCoOp \cite{yao2023visual} 
         & \small \color{gray}{58.2} & \scriptsize -
         & \small 52.6 & \scriptsize - 
         \\
         
         & \small CIP Reg 
         & \small \color{gray}{57.5}  & \scriptsize \color{gray}{\textit{-1.2}} 
         & \small 53.5 & \scriptsize \color{Greenish}{+0.9}
         \\
         
         & \small LLM Reg & \small \color{gray}{58.5}  & \scriptsize \color{gray}{\textit{-0.2}}
         & \small 53.1 & \scriptsize \color{Greenish}{+0.5} 
         \\
         
         & \small CIP+LLM Reg  
         & \small \color{gray}{57.6} & \scriptsize \color{gray}{\textit{-1.1}} 
         & \small \textbf{53.9} & \scriptsize \color{Greenish}{+1.3} 
         \\
         
    \hline\hline
        
       \rowcolor{LLGray} \cellcolor{white} RN50 
        
         & \small CoOp \cite{zhou2022learning} 
        & \small \color{gray}{51.4} & \scriptsize -
        & \small 45.6 & \scriptsize -
        \\
        
        \rowcolor{LLGray} \cellcolor{white}  & \small CoCoOp \cite{Zhou_2022_CVPR} 
    & \small \color{gray}{51.1} & \scriptsize - 
        & \small 46.3 & \scriptsize - 
 \\
        
        \rowcolor{LLGray} \cellcolor{white} & \small KgCoOp \cite{yao2023visual} 
        & \small \color{gray}{\textbf{51.8}} & \scriptsize - 
        & \small 46.9 & \scriptsize - 
        \\
        
        & \small CIPReg  
        & \small \color{gray}{50.6} & \scriptsize \color{gray}{\textit{-1.2}} 
        & \small 47.6 & \scriptsize \color{Greenish}{+0.7} 
        \\
        
         & \small LLMReg 
         & \small \color{gray}{\textbf{51.8}}   & \scriptsize \color{gray}{\textit{0.0}} 
         & \small 47.4 & \scriptsize \color{Greenish}{+0.5} 
         \\
         
         & \small CIP+LLMReg    
         & \small \color{gray}{51.1} & \scriptsize \color{gray}{\textit{-0.7}} 
         & \small \textbf{48.3} & \scriptsize \color{Greenish}{+1.4} 
         \\

    \hline
    \end{tabular}
    \end{center}
    \vspace{-5mm}
    \caption{\textbf{Regularizing soft prompts with geographical knowledge, top-1 bal. accuracy on GeoNet.} The regularization method accomplishes our goal to increase \textit{target} performance in GeoNet's USA-to-Asia transfer setting. CIP = CountryInPrompt, LLM = CountryLLM, CIP+LLM = CountryInPrompt+LLM. }
    \label{table:soft_prompt_2}
\end{table}
 \setlength{\tabcolsep}{2.0pt}
   
\begin{table}
    \begin{center}
    \renewcommand{\arraystretch}{0.8}

    \begin{tabular}{c|cc|cc|cc|cc}
     \multicolumn{1}{c|}{} & \multicolumn{8}{c}{\small \textbf{Threshold \textit{t} (\# Classes)}} \\
     \cline{2-9}

     \small \textbf{Method}
     & \multicolumn{2}{c|}{\small \textbf{$<$40\%}}
     & \multicolumn{2}{c|}{\small \textbf{$<$60\%}}
     & \multicolumn{2}{c|}{\small \textbf{$<$80\%}} 
     & \multicolumn{2}{c}{\small \textbf{$\leq$100\%}}  \\
     
     & (13) & \small $\Delta$ 
     & (45) & \small $\Delta$ 
     & (77) & \small $\Delta$ 
     & (95) & \small $\Delta$ \\
    \hline
    \hline
    \rowcolor{LLGray} \small CoOp \cite{zhou2022learning}
    & \small 31.2 & - 
    & \small 45.6 & - 
    & \small 55.6 & - 
    & \small 61.7 & - \\

    \rowcolor{LLGray} \small CoCoOp \cite{Zhou_2022_CVPR}
    & \small 32.8 & \scriptsize \color{Greenish}{+1.6} 
    & \small 45.4 & \scriptsize \color{red}{-0.2} 
    & \small 55.2 & \scriptsize \color{red}{-0.4} 
    & \small 61.4 & \scriptsize \color{red}{-0.3} \\
    
    \rowcolor{LLGray} \small KgCoOp  
    \cite{yao2023visual} 
    & \small 35.3 & \scriptsize \scriptsize \color{Greenish}{+4.1}
    & \small 47.9 & \scriptsize \scriptsize \color{Greenish}{+2.3}
    & \small 56.7 & \scriptsize \scriptsize \color{Greenish}{+1.1} 
    & \small 62.4 & \scriptsize \scriptsize \color{Greenish}{+0.7} \\
    \small CIPReg  
    & \small 37.5 & \scriptsize \color{Greenish}{+6.3} 
    & \small 48.9 & \scriptsize \color{Greenish}{+3.3} 
    & \small 57.8 & \scriptsize \color{Greenish}{+2.2} 
    & \small 63.5 & \scriptsize \color{Greenish}{+1.8} \\
    \small LLMReg   
    & \small 36.8 & \scriptsize \color{Greenish}{+5.6} 
    & \small 48.1 & \scriptsize \color{Greenish}{+2.5} 
    & \small 57.2 & \scriptsize \color{Greenish}{+1.6} 
    & \small 63.2 & \scriptsize \color{Greenish}{+1.5} \\
    \small CIP+LLMReg  
    & \small \textbf{39.9} & \scriptsize \color{Greenish}{+8.7} 
    & \small \textbf{49.5} & \scriptsize \color{Greenish}{+3.9} 
    & \small \textbf{58.2} & \scriptsize \color{Greenish}{+2.6} 
    & \small \textbf{64.0} & \scriptsize \color{Greenish}{+2.3} \\
    \hline
    \end{tabular}
    \end{center} 
    \vspace{-4mm}
        \caption{\textbf{Performance on DollarStreet classes with less than \textit{t}\% recall in CoOp}, with ViT-B/16. Gains w.r.t. CoOp of our geography knowledge regularization are especially large for CoOp's difficult classes (+8.7 in $<$40\%), compared to KgCoOp's (+4.1 in $<$40\%, i.e. a \textbf{4.6} difference from ours). CIP = CountryInPrompt, LLM = CountryLLM, CIP+LLM = CountryInPrompt+LLM.}
    \label{table:per_class_perf}
\end{table}
\begin{figure}[t]
    \centering
    \includegraphics[scale=0.135]
    {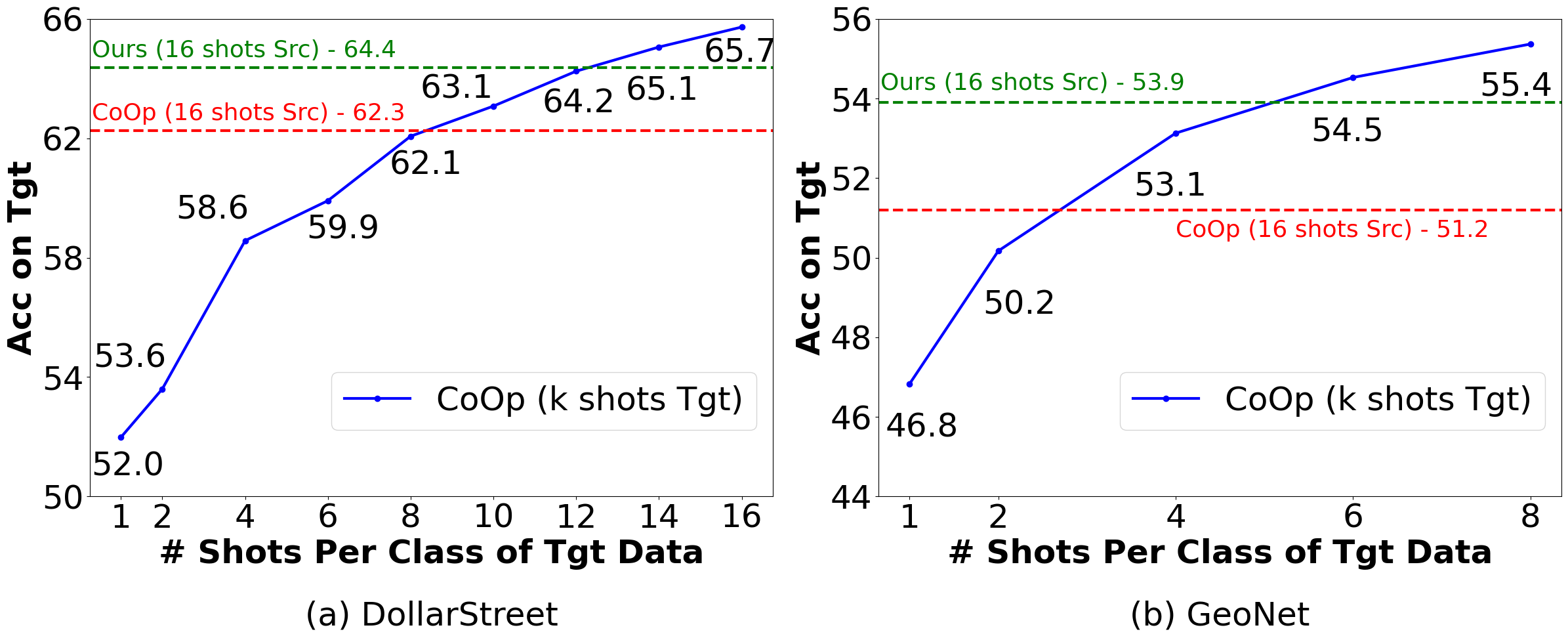}
    \vspace{-2mm}
    \caption{\textbf{Geography knowledge-regularized soft prompts trained on source data (ours, \textcolor{Greenish}{green} line) vs. few-shot soft prompts trained on target data (\textcolor{blue}{blue} curve)}. (a) Src=Europe, Tgt=Africa,Asia,Amer.; (b) Src=USA,Tgt=Asia. Our 16-shot model trained on only source data (\textcolor{Greenish}{green}) outperforms a model with prompts trained on 12 or 4 shots \emph{per class} of target data (on DollarStreet\&GeoNet, resp.), which is 1140\&2400 images total.} 
    \label{few_shot_tgt_fig}
\end{figure}

\noindent \textbf{Regularization helps significantly on difficult classes.} As certain objects may be especially sensitive to geographical domain shift, we break down classwise performance on DollarStreet in Table \ref{table:per_class_perf}, using a stratification of class difficulty based on default soft prompting performance (CoOp). \textit{The CountryInPrompt+LLM strategy achieves significant gains on the classes most difficult with respect to the CoOp baseline}. In particular, gains of +8.7\% in balanced accuracy are achieved for classes with $<$40\% baseline recall, while the highest achieved by KgCoOp is 4.1\%. Example classes in this subset are \textit{snacks}, \textit{clothes}, and \textit{makeup}. The DollarStreet classes with greatest improvement, independent of original CoOp accuracy are: \textit{piercings}, \textit{clothes}, \textit{homes}, \textit{medication}, and \textit{refrigerators} (all at least +14\% over CoOp). In GeoNet, 
\textit{dome}, \textit{goby} (fish), \textit{eland} (antelope), and \textit{gloriosa} (flower) have $>$20 samples and $>$30\% improvement. A total of 64/95 classes in DollarStreet and 209/344 in GeoNet improve vs. CoOp, showing broad coverage. 
        
\noindent \textbf{Regularized source-only prompts outperform few-shot target-trained prompts.} Given that soft prompts can show strong performance in few-shot settings, a potential alternative to regularizing soft prompts on source data is to directly acquire a few examples of target data for training. We evaluate this setting by splitting target data into train/test, and training CoOp at varying \# of shots of \textit{target} data for GeoNet and DollarStreet, shown in Fig.~\ref{few_shot_tgt_fig}. Notably, training on 16 shots per class of \emph{source} data with \emph{our regularization method} outperforms using 12 \& 4 shots per class of \emph{target} data on DollarStreet \& GeoNet. This is a vast amount of target data overall (\eg 12 shots x 95 classes = 1140 target samples in DollarStreet, 4 shots x 600 classes = 2400 samples in GeoNet). The baseline CoOp trained on 16 shots of source data only outperforms an 8-shot/2-shot target-trained CoOp model (DollarStreet/GeoNet). Our strategy is thus more compelling in the absence of a lot of target data. 

\noindent \textbf{Performance by income.} DollarStreet provides  
estimated monthly income of the household in which an image was captured. We evaluate with the delineation of low, medium, and high-income buckets from \cite{goyal2022fairness}.
Compared to CoOp/KgCoOp, CountryInPrompt+LLM gains are +2.5/+3.4 in low, +2.4/+1.5 in medium, and +2.1/+0.7 in high. Thus our method especially improves in low-income areas, though it helps across levels. The table is in supp. 


\subsection{Further Analysis}

\noindent \textbf{Are descriptions correlated with key country statistics?} 
For a pair of countries, we compute two values. We measure the distance between each class embedding and take the average overall distance as one value. We also take the absolute difference between statistics for those countries (from \cite{wikipedia,worldbank_indicator}, \eg difference in avg. yearly temperature) as the other value. We compute the correlation between these two values over every unique country pair in DollarStreet, showing results in Table \ref{table:correl}. We find that the strongest correlation across each prompt type is with HDI, which summarizes human development. It is notable that factors like yearly temperature and precipitation also show moderate correlations, indicating a potential role of climate. Future work may further explore how object differences present with respect to these factors. It will also be critical to ensure that differences between countries are representative and not exaggerated in embeddings.
        
 \setlength{\tabcolsep}{1pt}

\begin{table}[t]

    \begin{center}
        \renewcommand{\arraystretch}{0.8}

    \begin{tabular}{c|c|c|c}
        \hline
         \small \textbf{Statistic} & \small \textbf{CIP} & \small  \textbf{CountryLLM} & \small 
            \textbf{CIP+LLM}
         \\
          \hline
         \small GDP Per Capita (US \$) & \small \textbf{0.219} & \small 0.063 & \small \textbf{0.217}  \\
          \small Human Devel. Index (HDI) & \small \textbf{0.439} & \small \textbf{0.385} & \small \textbf{0.451}  \\
          \small Land Area (km$^2$) & \small -0.072 & \small 0.050* & \small -0.046*  \\
          \small Population (\#) & \small -0.131 & \small 0.077 & \small -0.123  \\ 
          \small Population Density (\#/km$^2$) & \small 0.103 & \small 0.158  & \small 0.081  \\ 
          \small \% Agricultural Land & \small 0.139 & \small 0.070 & \small 0.122  \\
          \small \% Forest Area & \small 0.191 & \small 0.087 & \small \textbf{0.201}  \\
          \small Avg. Yearly Temp. ($^\circ$C) & \small \textbf{0.380} & \small \textbf{0.256} & \small \textbf{0.391} \\
          \small Avg. Yearly Precip. (mm/year) & \small \textbf{0.236} & \small 0.124 & \small \textbf{0.230}  \\

    \hline
    \end{tabular}
    \end{center}
    \vspace{-5mm}
    \caption{\textbf{Correlation (Pearson's $\rho$) of avg. CLIP class text embedding distance and country statistic difference}, \eg economic (GDP per capita, HDI) and climate factors (temperature, precipitation, forest area). We use the 63 countries in DollarStreet (1,953 pairs). \textbf{Bold} values have $\rho$$>$0.2, * means not significant ($\alpha$=0.01).}
    \label{table:correl}
\end{table}

\begin{figure}
    \includegraphics[scale=0.215]{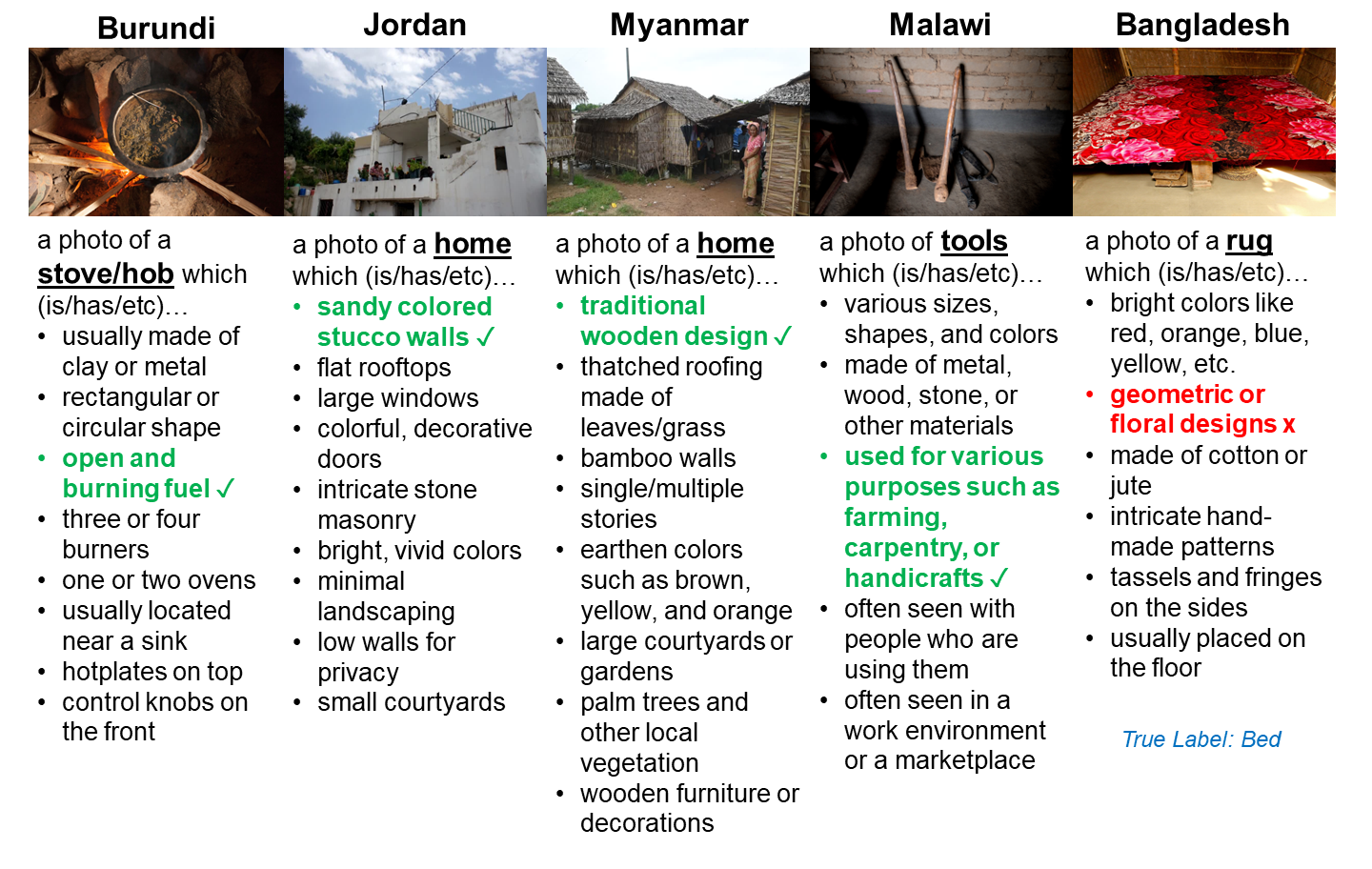}
    \vspace{-4mm}
    \caption{\textbf{Qualitative analysis.} We show examples where geography-specific descriptors improve/hurt vs. general descriptors in zero-shot inference. We highlight the prediction's descriptors, bolding the highest activating one. Encoder=RN50.}
    \label{fig:qual}
\end{figure}

\begin{figure}
\centering
    \includegraphics[width=0.78\linewidth]{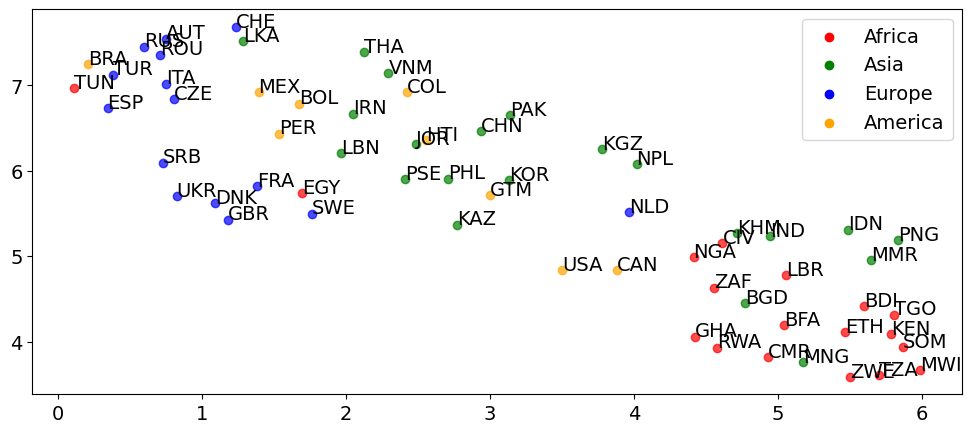}
    \caption{\textbf{UMAP \cite{mcinnes2018umap-software} plot for CountryLLM and the category \textit{homes} in DollarStreet.} Country-specific descriptors are often close to those of other countries intra-continent, likely due to similar weather, environment, and/or economic conditions.}
    \label{fig:umap}
\end{figure}

\noindent \textbf{Descriptor topics.} 
We show a UMAP \cite{mcinnes2018umap-software} visualization comparing CountryLLM text embeddings for the category \textit{homes} across geographies in Fig. \ref{fig:umap}. Countries tend to group by continent, showing the representations may capture similarity in features like climate and/or economics. We examine a few topics mentioned in the CountryLLM descriptors for \textit{homes}. While ``stone'' is described across continents, ``bright colors'' and ``mud'' are mentioned mostly in Africa, and ``balcony'' in Europe and Asia. We show more in supp. 

\noindent \textbf{Success and failure examples.} 
We provide examples of CountryLLM vs. GeneralLLM in Fig. \ref{fig:qual} on DollarStreet. The model captures geographical descriptive knowledge like ``sandy colored stucco walls'' for \textit{homes} in Jordan, a feature which may be less common for Western homes. Sometimes the model may be too attentive to attributes, leading to confusion (\eg choosing \textit{rug} over \textit{bed}). Future work that enhances alignment in VLMs can likely improve results.

\section{Conclusion}
\label{sec:conc}


In this work, we bring attention to how various strategies to prompt CLIP affect recognition performance across geographies. In addition, through soft prompting with descriptive knowledge, we provide a mechanism to achieve a more geo-generalizable set of class representations across regions. Our work is only a first step in this important area.

\noindent \textbf{Limitations and ethical considerations.} While our method's proof of concept is demonstrated in a positive effort to debias CLIP's default representations through diversity, due to the biased worldview of the Internet, CLIP's representations are likely inadequate, exaggerated, and/or not fully representative for some countries. While we expect quality LLM knowledge to guide better representations, LLM knowledge can also be incorrect (e.g., through hallucination), imprecise, or biased.

\noindent \textbf{Future work.} For the above reasons, our future efforts aim to ensure more representative VLM/LLM knowledge.
We strongly advocate for the community to seek communication with diverse groups within all countries (i.e. to capture areas that range from low to high income) to ensure better representation and fairness in AI technology use. There are notable continent-level disparities to still improve upon. 

\noindent \textbf{Acknowledgement.} This work was supported by National Science Foundation Grants No. 2006885 and 2329992.
{
    \small
    \bibliographystyle{ieeenat_fullname}
    \bibliography{main}
}

\clearpage

    \section{Supplementary Material}

    The supplementary material is organized as extensions to Sec 5.1 through 5.3 in the main paper (zero-shot inference, soft prompting, and further analysis). We provide in-depth experiments and ablations to support our main findings and method design. We also provide additional examples.

    \subsection{Zero-Shot Inference}

\begin{figure}[t]
    \centering
    \includegraphics[scale=0.118]
    {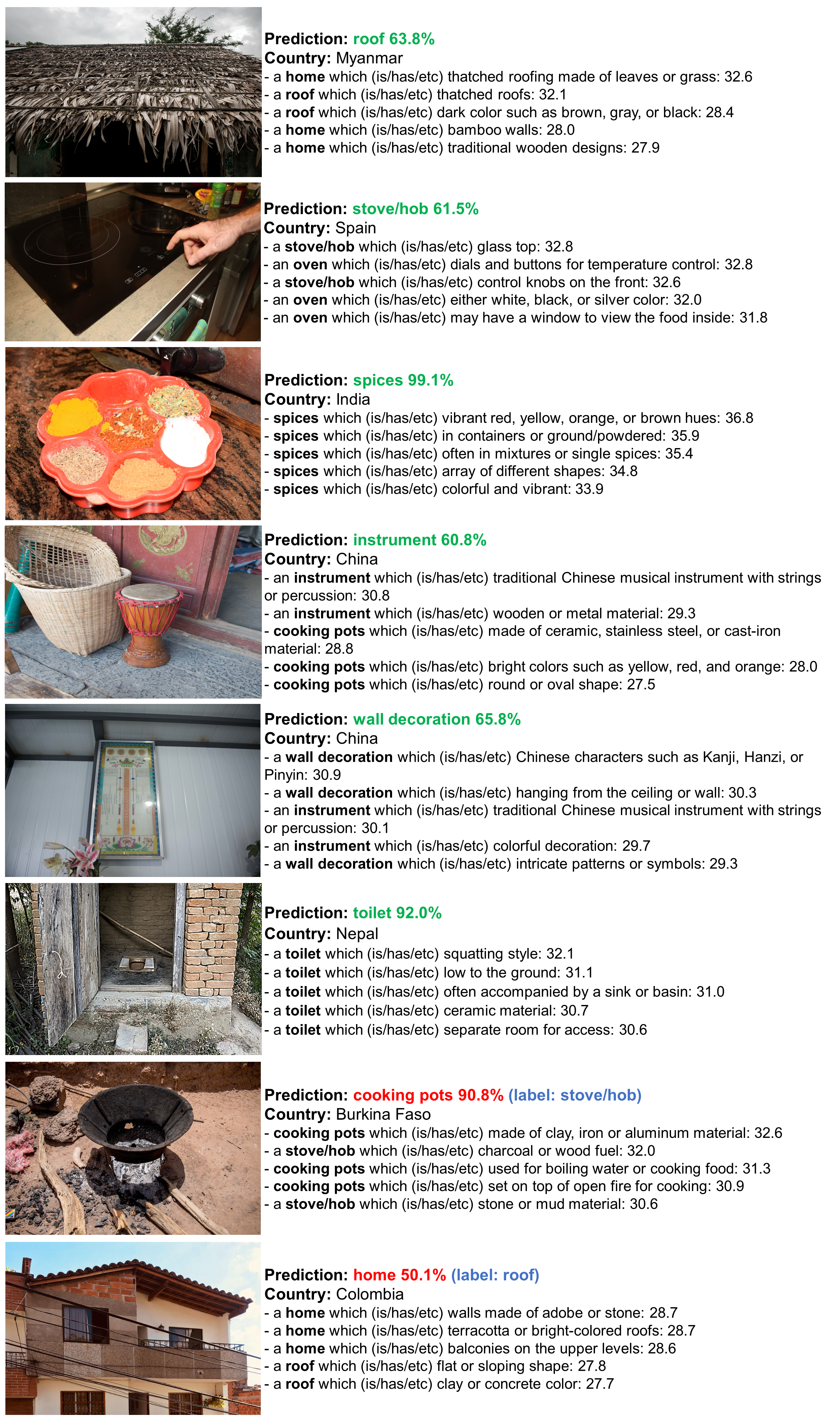}
    \vspace{-3mm}
    \caption{\textbf{Qualitative examples of success/failure cases (CountryLLM).} We show the prediction (\color{Greenish}{green}\color{black}{} if correct, \color{red}{red}\color{black}{} if not) as well as the prediction confidence and the top 5 descriptors (with CLIP similarity scores) for each image. Encoder = ViT-B/16.} 
    \label{supp:qual}
\end{figure}

           \noindent \textbf{Qualitative examples.} In Fig. \ref{supp:qual}, we provide additional examples of zero-shot CLIP inference on DollarStreet using geography-specific descriptors (CountryLLM). It is demonstrated that geography knowledge is successfully able to capture diverse object forms and designs. The top activating descriptors in these examples highlight materials (``thatch'' for \textit{roof} in Myanmar, ``glass'' for \textit{stove/hob} in Spain) and colors (``yellow''/``orange'' for \textit{spices} in India). LLM context enables CLIP to be probed for its own cultural knowledge (\eg ``traditional Chinese musical instrument'' for \textit{instrument} and ``Chinese characters such as Kanji, Hanzi, or Pinyin'' for \textit{wall decoration}). Cultural conventions are better captured, exhibited by ``squatting style'' activating for \textit{toilet} in Nepal (such form is not common in Western geographies). In error cases, some classes with related descriptors may be confused (\textit{cooking pots} and \textit{stove/hob}). Such errors suggest improvement is needed in CLIP's understanding of natural language concepts. Error cases may also result because of ambiguity with close categories, as shown by \textit{home} vs. \textit{roof} in Colombia (where the \textit{home} descriptors seem fairly accurate). Nonetheless, it is interesting to observe that a successful prediction can occur even when descriptors from other categories strongly activate. 
            \begin{figure*}[t]
    \centering
    \includegraphics[scale=0.0525]
    {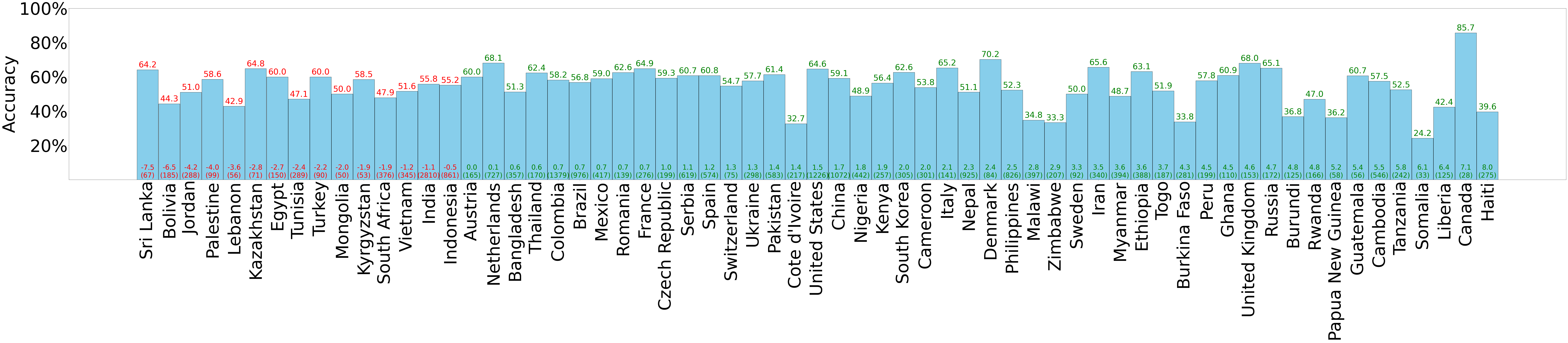}
    \vspace{-3mm}
    \caption{\textbf{Country-level overall accuracy in zero-shot inference with CountryInPrompt+LLM, \color{Greenish}{gains}\color{black}/\color{red}{drops}\color{black}{} shown vs. GeneralLLM (descriptors not specific to geographies)}, for ViT-B/16. Note that geography knowledge integration is generally effective across countries, demonstrated by performance improvements in 48/63 countries. The overall accuracy over all countries is 55.8\% for CountryInPrompt+LLM and 54.6\% for GeneralLLM.} 
    \label{supp:per_country}
\end{figure*}
             \setlength{\tabcolsep}{1pt}

\begin{table*}[t]

    \begin{center}
    \renewcommand{\arraystretch}{0.8}

    \begin{tabular}{c|c||cc|gg|cc|gg}
        
        \hline 
        & & \multicolumn{4}{c||}{\textbf{\textit{\small Top-1 Accuracy}}} 
        & \multicolumn{4}{c}{\textbf{\textit{\small Top-3 Accuracy}}} 
        \\
        
        \textbf{\small Encoder} & \textbf{\small Prompting Method} & \multicolumn{2}{c}
        {\textbf{\small USA}} & \multicolumn{2}{c||}{\textbf{\small Asia}} &
        \multicolumn{2}{c}{\textbf{\small USA}} & \multicolumn{2}{c}{\textbf{\small Asia}}  \\
         & 
         &  \multicolumn{1}{c}{\small Acc} 
         & \multicolumn{1}{c}{\small $\Delta$} 

         & \multicolumn{1}{c}{\small Acc} 
         & \multicolumn{1}{c||}{\small $\Delta$} 

             &  \multicolumn{1}{c}{\small Acc} 
         & \multicolumn{1}{c}{\small $\Delta$} 


         & \multicolumn{1}{c}{\small Acc} 
         & \multicolumn{1}{c}{\small $\Delta$} 
         \\
         
    \hline\hline \rowcolor{LLGray}
        \cellcolor{white} \small ViT-B/32  
        & \small Zero-Shot CLIP \cite{radford2021learning} 
        & \small 50.3 & \scriptsize - 
        & \small 46.2 & \scriptsize - 
        & \small 69.6 & \scriptsize - 
        & \small 66.1 & \scriptsize -

        \\
        
         \rowcolor{LLGray}
        \small \cellcolor{white} & \small GeneralLLM \cite{menon2022visual}
         & \small 49.9 & \scriptsize \color{red}{-0.4} 
         & \small 46.6 & \scriptsize \color{Greenish}{+0.4} 
         & \small 69.1 & \scriptsize \color{red}{-0.5} 
        & \small 67.0 & \scriptsize \color{Greenish}{+0.9}
      
         \\

         &  \small CountryInPrompt 

         & \small 50.0 & \scriptsize \color{red}{-0.3} 
         & \small 46.5 & \scriptsize \color{Greenish}{+0.3} 
         & \small 69.1 & \scriptsize \color{red}{-0.5} 
        & \small 66.3 & \scriptsize \color{Greenish}{+0.2}
         \\
         
         &  \small CountryLLM 

        & \small 50.3 & \scriptsize \color{black}{0.0} 
         & \small 47.7 & \scriptsize \color{Greenish}{+1.5} 
         & \small 69.7 & \scriptsize \color{Greenish}{+0.1} 
        & \small 67.1 & \scriptsize \color{Greenish}{+1.0}
        \\

         & \small CountryInPrompt+LLM  
        & \small \textbf{51.2} & \scriptsize \color{Greenish}{+0.9} 
         & \small \textbf{47.8} & \scriptsize \color{Greenish}{+1.6} 
         & \small \textbf{69.8} & \scriptsize \color{Greenish}{+0.2} 
        & \small \textbf{67.7} & \scriptsize \color{Greenish}{+1.6}
        \\
         
    \hline\hline \rowcolor{LLGray}
         \cellcolor{white} \small ViT-B/16  &  \small Zero-Shot CLIP \cite{radford2021learning}  
         & \small 53.9 & - 
         & \small 50.2 & -
         & \small 72.8 & -
        & \small 69.2 & -
       \\
       
         \rowcolor{LLGray}
        \cellcolor{white} & \small GeneralLLM \cite{menon2022visual} 
         & \small 54.6 & \scriptsize \color{Greenish}{+0.7} 
         & \small 52.2 & \scriptsize \color{Greenish}{+2.0} 
         & \small 73.4 & \scriptsize \color{Greenish}{+0.6} 
        & \small 70.9 & \scriptsize \color{Greenish}{+1.7}

        \\

         & \small CountryInPrompt 
         & \small 54.6 & \scriptsize \color{Greenish}{+0.7} 
         & \small 50.7 & \scriptsize \color{Greenish}{+0.5} 
         & \small 73.3 & \scriptsize \color{Greenish}{+0.5} 
        & \small 70.5 & \scriptsize \color{Greenish}{+1.3}

         \\

         & \small  CountryLLM  
         & \small 54.7 & \scriptsize \color{Greenish}{+0.8} 
         & \small \textbf{52.5} & \scriptsize \color{Greenish}{+2.3} 
         & \small 73.5 & \scriptsize \color{Greenish}{+0.7} 
        & \small \textbf{71.1} & \scriptsize \color{Greenish}{+1.9}
         \\
         
         &  \small CountryInPrompt+LLM 
             & \small \textbf{54.9} & \scriptsize \color{Greenish}{+1.0} 
         & \small 51.4 & \scriptsize \color{Greenish}{+1.2} 
         & \small \textbf{74.1} & \scriptsize \color{Greenish}{+1.3} 
        & \small 70.9 & \scriptsize \color{Greenish}{+1.7}

         \\
    \hline
    \hline
    \rowcolor{LLGray}
         \cellcolor{white} \small RN50 
        & \small Zero-Shot CLIP \cite{radford2021learning} 
         & \small 46.8 & -
         & \small 43.4 & -
         & \small 65.6 & -
        & \small 63.2 & -
       \\

         \rowcolor{LLGray}
        \small \cellcolor{white} & \small GeneralLLM \cite{menon2022visual}
         & \small 48.6 & \scriptsize \color{Greenish}{+1.8} 
         & \small 45.4 & \scriptsize \color{Greenish}{+2.0} 
         & \small 67.3 & \scriptsize \color{Greenish}{+1.7} 
        & \small 65.5 & \scriptsize \color{Greenish}{+2.3}
         \\

         & \small CountryInPrompt 

         & \small 47.5 & \scriptsize \color{Greenish}{+0.7} 
         & \small 43.9 & \scriptsize \color{Greenish}{+0.5} 
         & \small 66.8 & \scriptsize \color{Greenish}{+1.2} 
        & \small 63.7 & \scriptsize \color{Greenish}{+0.5}
        \\

         &  \small CountryLLM 
         & \small 48.2 & \scriptsize \color{Greenish}{+1.4} 
         & \small \textbf{45.7} & \scriptsize \color{Greenish}{+2.3} 
         & \small 67.3 & \scriptsize \color{Greenish}{+1.7} 
        & \small 65.2 & \scriptsize \color{Greenish}{+2.0}
        \\
         
         &  \small CountryInPrompt+LLM 
            & \small \textbf{49.2} & \scriptsize \color{Greenish}{+2.4} 
         & \small 45.4 & \scriptsize \color{Greenish}{+2.0} 
         & \small \textbf{67.8} & \scriptsize \color{Greenish}{+2.2} 
        & \small \textbf{65.7} & \scriptsize \color{Greenish}{+2.5}
         \\ 
    \hline
    \end{tabular}
    \end{center}
    \vspace{-5mm}
    \caption{\textbf{Zero-shot CLIP with descriptive knowledge prompts, top-1/3 balanced accuracy (Acc) on GeoNet.} Strategies to capture CLIP's internal country knowledge (CountryInPrompt), external LLM country knowledge (CountryLLM), and their combination (CountryInPrompt+LLM), improve the zero-shot CLIP baseline (prompt ``a photo of a''). Gains in \color{Greenish}{green}\color{black}, drops in \color{red}{red}\color{black}. }
    \label{table:supp_zero_shot_geonet}
\end{table*}

        \noindent \textbf{DollarStreet performance by country.} The zero-shot, continent-level DollarStreet results in Table 1 of the main paper can be further broken down into country-level performance. We particularly show CountryInPrompt+LLM vs. GeneralLLM (i.e. full geography knowledge vs. general knowledge) with ViT-B/16 in Fig. \ref{supp:per_country}. This figure notably exhibits per-country \textit{overall} accuracy instead of balanced accuracy due to limited examples per class per country. Here we see that compared to GeneralLLM, 48/63 countries have performance improvements with geography knowledge. 15 countries show drops, for which we posit a couple of reasons. For one, CLIP may have inadequate understanding of some detailed LLM descriptors due to imperfect vision-language alignment. Secondly, it is possible that the geography knowledge may be insufficient for some countries. Ensuring proper alignment between visual and language features and more adequate quality of LLM+VLM knowledge are important areas for future work.

        \noindent \textbf{GeoNet.} We further show zero-shot inference with CLIP on GeoNet (test sets) in Table \ref{table:supp_zero_shot_geonet}. The 10 most common countries in the ``Asia'' test set are used for our ``Asia'' evaluation to match the geography-specific LLM descriptors we acquire (i.e. China, India, Indonesia, Japan, Kenya, Malaysia, Singapore, Taiwan, Tanzania, and Thailand - GeoNet considers these as all ``Asia''). Observe that the highest USA/Asia performance for each encoder is provided by one of the geography-specific prompting strategies. With ViT-B/32, CountryLLM has notably higher performance on Asia vs. GeneralLLM. With ViT-B/16 and RN50,  CountryLLM improves vs. GeneralLLM on Asia, though top target differences are smaller compared to DollarStreet. We hypothesize that the general concept representations probed by LLMs and the ones respective to the top countries in Asia are relatively similar within GeoNet (\eg ~43\% of images are from Japan/China, which are high-resource). In general, the default CLIP gaps between USA and Asia are not extremely large, indicating that CLIP has a notable degree of robustness on GeoNet. Also, unlike DollarStreet, some classes within GeoNet are mostly unique to geographies (\eg \textit{shoji} in traditional Japanese architecture), so cross-geography knowledge may not be as helpful. 
        

    \subsection{Soft Prompting}

        \noindent \textbf{Performance by income.} Instead of reporting target performance by continent, in Table \ref{table:supp_income} we show results breaking down target performance by income (using the delineation of low, medium, and high-income buckets from \cite{goyal2022fairness}). While our method helps across all income levels, the gains are most significant in low-income scenarios.

        \noindent \textbf{Ablation: regularization weight.} In Table \ref{table:supp_reg_weight}, we show CountryInPrompt+LLM performance for various choices of $\lambda$. The highest total performance is achieved at $\lambda=4$.

        \noindent \textbf{Experiment: source of knowledge.} With CountryInPrompt+LLM, we test three different ways to select countries for knowledge aggregation (i.e. how to choose $\mathcal{G}_t$): (1) using unseen countries of interest (named \textit{target}, with country count 49), (2) using countries seen during training (named \textit{source}, with country count 14), and (3) using all countries in the dataset (named \textit{all}, with country count 63). 
        Shown in Table \ref{table:supp_source_knowl}, we find that both \textit{target} and \textit{all} methods perform well on target geographies in comparison to source, and the target-only ensemble does best overall. This result indicates that including diverse knowledge of target countries best ensures geographical robustness across the world. With RN50, using a source-only ensemble performs poorly on Africa and Asia, but best on Americas (presumably due to some greater similarities, \eg between the US, Canada, and Europe). Interestingly, we find that target regularization is best on the source test set, which we attribute to more domain-generalizable class representations achieved overall, given the use of diverse knowledge. 

             \setlength{\tabcolsep}{0.5pt}

\begin{table}[t]
    \begin{center}
        \renewcommand{\arraystretch}{0.8}

    \begin{tabular}{c|cc|cc|cc|cc}
        \hline
         & \multicolumn{2}{c|}{\textit{\textbf{Source}}} 
         & \multicolumn{6}{c}{\textit{\textbf{Target}} \textbf{(Income Status)}}   \\
        
         \textbf{Method} & \multicolumn{2}{c|}{\textbf{Europe}} &\multicolumn{2}{c}{\textbf{Low}} &\multicolumn{2}{c}{\textbf{Medium}} &\multicolumn{2}{c}{\textbf{High}} \\
        
          & \multicolumn{1}{c}{\small Acc} & \multicolumn{1}{c|}{\small $\Delta$} & 
         \multicolumn{1}{c}{\small Acc} & \multicolumn{1}{c}{\small $\Delta$}  

          &  \multicolumn{1}{c}{\small Acc} & \multicolumn{1}{c}{\small $\Delta$} & 
         \multicolumn{1}{c}{\small Acc} & \multicolumn{1}{c}{\small $\Delta$} 
         
         \\
    
    \hline\hline
    
         \rowcolor{LLGray} 
        
          \small CoOp \cite{zhou2022learning} 
         & \small \small \color{gray}{72.2} & \scriptsize -  
         & \small 44.3 & \scriptsize - 
         & \small 61.6 & \scriptsize -  
         & \small 71.1 & \scriptsize - 
         \\

         \rowcolor{LLGray} \small CoCoOp \cite{Zhou_2022_CVPR} 
         & \small \small \color{gray}{73.2} & \scriptsize -  
         & \small 44.4 & \scriptsize - 
         & \small 61.4 & \scriptsize -  
         & \small 70.5 & \scriptsize - 
         \\

         \rowcolor{LLGray} \small KgCoOp \cite{yao2023visual} 
         & \small \small \color{gray}{73.1} & \scriptsize -  
         & \small 43.4 & \scriptsize - 
         & \small 62.5 & \scriptsize -  
         & \small 72.6 & \scriptsize - 
         \\
         
          \small CIP Reg 
         & \small \color{gray}{71.8} & \scriptsize \textit{\color{gray}{-1.4}}   
         & \small 46.0 & \scriptsize \color{Greenish}{+1.6}
         & \small 63.6 & \scriptsize \color{Greenish}{+1.1}
         & \small 72.4 & \scriptsize \color{red}{-0.2}
         \\
         
          \small LLM Reg &
         \small \color{gray}{73.2} & \scriptsize \textit{\color{gray}{0.0}}  
         & \small 45.2 & \scriptsize \color{Greenish}{+0.8}
         & \small 63.1 & \scriptsize \color{Greenish}{+0.6}
         & \small \textbf{73.3} & \scriptsize \color{Greenish}{+0.7}
         \\
         
          \small CIP+LLM Reg  
         & \small \color{gray}{\textbf{73.6}} & \scriptsize \textit{\color{gray}{+0.4}} 
         & \small \textbf{46.8} & \scriptsize \color{Greenish}{+2.4}
         & \small \textbf{64.0} & \scriptsize \color{Greenish}{+1.5}
         & \small \textbf{73.3} & \scriptsize \color{Greenish}{+0.7}
         \\

    \hline
    \end{tabular}
    \end{center}
    \vspace{-5mm}
    \caption{\textbf{Regularizing soft prompts with geography knowledge, top-1 bal. accuracy on DollarStreet, with target organized by income status.} Gains w.r.t. best baseline. Note that geo-diverse prompts help especially in the low-income scenario. CIP = CountryInPrompt, LLM = CountryLLM, CIP+LLM= CountryInPrompt+LLM. Encoder = ViT-B/16.}
    \label{table:supp_income}
\end{table}
         \setlength{\tabcolsep}{2pt}

\begin{table}[t]
    \begin{center}
        \renewcommand{\arraystretch}{0.8}

    \begin{tabular}{c|c|c|c|c|c}
        \hline
         & \multicolumn{1}{c|}{\textit{\textbf{Source}}} 
         & \multicolumn{4}{c}{\textit{\textbf{Target}}}   \\
        
         \textbf{\small $\lambda$} & \multicolumn{1}{c|}{\textbf{Europe}} &\multicolumn{1}{c}{\textbf{Africa}} &\multicolumn{1}{c}{\textbf{Asia}} &\multicolumn{1}{c}{\textbf{Americas}} &\multicolumn{1}{c}{\textbf{Total}} \\
        
          & \multicolumn{1}{c|}{\small Acc}  & 
         \multicolumn{1}{c}{\small Acc} &  \multicolumn{1}{c}{\small Acc} & 
         \multicolumn{1}{c}{\small Acc}
           &  \multicolumn{1}{c}{\small Acc} 
 
         \\
    
    \hline\hline

          \small 2
         & \small \small \color{gray}{73.1}  
         & \small 55.9 
         & \small 63.0  
         & \small \textbf{70.3}
          & \small 63.4  
         \\

         \small 4
         & \small \small \textbf{\color{gray}{73.6}}
         & \small 57.2  
         & \small \textbf{63.8} 
         & \small \textbf{70.3} 
          & \small \textbf{64.0}  
         \\

         \small 6
         & \small \small \color{gray}{72.8}
         & \small \textbf{57.3}
         & \small 63.5  
         & \small \textbf{70.3} 
          & \small 63.8 
         \\
         
         \small 8
         & \small \color{gray}{72.7}
         & \small 56.7 
         & \small 63.1 
         & \small 68.9
          & \small 63.1 
         \\
         
         \small 10 &
         \small \color{gray}{70.7}   
         & \small 55.4 
         & \small 61.6 
         & \small 68.1 
          & \small 62.0  
         \\

    \hline
    \end{tabular}
    \end{center}
    \vspace{-5mm}
    \caption{\textbf{Regularizing soft prompts with geography knowledge, top-1 bal. accuracy on DollarStreet, at varying values of $\lambda$.} 
    Our method uses $\lambda$=4. Encoder = ViT-B/16.}
    \label{table:supp_reg_weight}
\end{table}
        \setlength{\tabcolsep}{2pt}

\begin{table}[t]
    \begin{center}
        \renewcommand{\arraystretch}{0.9}

    \begin{tabular}{c|c|c|c|c|c|c}
        \hline
         &  & \multicolumn{1}{c|}{\textit{\textbf{Source}}} 
         & \multicolumn{4}{c}{\textit{\textbf{Target}}}   \\
        
         Encoder & \textbf{$\mathcal{G}_t$} & \multicolumn{1}{c|}{\textbf{Europe}} &\multicolumn{1}{c}{\textbf{Africa}} &\multicolumn{1}{c}{\textbf{Asia}} &\multicolumn{1}{c}{\textbf{Americas}} &\multicolumn{1}{c}{\textbf{Total}} \\
        
         & & \multicolumn{1}{c|}{\small Acc}  & 
         \multicolumn{1}{c}{\small Acc} &  \multicolumn{1}{c}{\small Acc} & 
         \multicolumn{1}{c}{\small Acc}
           &  \multicolumn{1}{c}{\small Acc} 
 
         \\
    
    \hline\hline

         \small ViT-B/16 & \small Target
         & \small \small \textbf{\color{gray}{73.6}}
         & \small \textbf{57.2}  
         & \small \textbf{63.8} 
         & \small \textbf{70.3} 
          & \small \textbf{64.0}  
         \\

         &  \small All
         & \small \small \color{gray}{72.6}  
         & \small 56.6 
         & \small 63.6 
         & \small 70.1
          & \small 63.8 \\

           &  \small Src
         & \small \color{gray}{73.5}
       & \small 55.8
        & \small 62.9
         & \small 70.1
          & \small 63.1

         \\
         \hline
         \small RN50 & \small Target
         & \small \small \textbf{\color{gray}{65.5}}
            & \small \textbf{48.1} 

         & \small \textbf{54.5} 
                 & \small 60.4 
          & \small \textbf{54.8}  
         \\

         &  \small All
         & \small \small \textbf{\color{gray}{65.5}} 
        & \small \small \textbf{48.1}  
             & \small \textbf{54.5} 
         & \small 60.3

            & \small \textbf{54.8}
          \\

      &  \small Src
        & \small \color{gray}{64.9}

         & \small 46.8
      & \small 54.4
         & \small \textbf{60.6}
                 & \small 54.5 \\


    \hline
    \end{tabular}
    \end{center}
    \vspace{-5mm}
    \caption{\textbf{Regularizing soft prompts with geography knowledge, top-1 bal. accuracy on DollarStreet, with different countries in $\mathcal{G}_t$.} Comparisons are shown for CountryInPrompt+LLM at $\lambda$=4. Using a target country-only ensemble (49 countries) performs slightly better than using all countries (63 countries).}
    \label{table:supp_source_knowl}
\end{table}

        \noindent \textbf{Comparison to \textit{gpt-3.5-turbo}.} In addition to \textit{davinci-003}, we test \textit{gpt-3.5-turbo} (ChatGPT) as an LLM knowledge source. \textit{gpt-3.5-turbo} notably needed significant prompt engineering to produce adequate descriptors. We use the following prompt for \textit{gpt-3.5-turbo}:   

            \begin{quote}
                \small Task: For an object/concept name and country name provided, very concisely provide up to 10 visual features that can distinguish that object in a photo taken in that specific country. The key is to make sure the descriptions capture an object’s key visual attributes and properties across the country. Examples include colors, textures, shapes, materials used, parts/components, common context/background, size, and possible designs/forms across the country. Consider common attributes specific to objects in that country and ensure descriptor diversity to represent regions with low socioeconomic status. \\
                These are strict output requirements:
                \\
                - Each description should be simple and interpretable by a child
                \\
                - Use only a few words per descriptor 
                \\
                - Start directly in the form of a bulleted list
                \\
                - The output should complete this sentence: ``A/an $<$object$>$ which (is/has/etc.)''
                \\
                - Be specific, qualifying with visual adjectives, and do not be vague or general at all
                \\
                 - Adjectives like ``unique''/``diverse''/``distinctive'' are not specific enough to help distinguish an object in a photo, so do not use them \\
                - Specific EX: ``red color''/``small size''/``wooden hand''
                \\
                Use this example as a reference... 
                \\ To tell there is a bathtub in a photo in Japan, the following visual features are helpful:
                \\ - short in length and deep 
                \\ - square shape 
                \\ - wooden, plastic, or steel material
                \\ - white or brown color 
                \\ - benches on side 
                \\ - next to shower  \\
                Now complete:\\
                To tell there is $<$object$>$ in a photo in $<$country$>$, the following visual features are helpful: 
            \end{quote}

         \setlength{\tabcolsep}{2pt}

\begin{table}[t]
    \begin{center}
        \renewcommand{\arraystretch}{0.9}

    \begin{tabular}{c|c|c|c|c|c|c}
        \hline
         &  & \multicolumn{1}{c|}{\textit{\textbf{Source}}} 
         & \multicolumn{4}{c}{\textit{\textbf{Target}}}   \\
        
         Encoder & LLM & \multicolumn{1}{c|}{\textbf{Europe}} &\multicolumn{1}{c}{\textbf{Africa}} &\multicolumn{1}{c}{\textbf{Asia}} &\multicolumn{1}{c}{\textbf{Amer.}} &\multicolumn{1}{c}{\textbf{Total}} \\
        
         & &  \multicolumn{1}{c|}{\small Acc}  & 
         \multicolumn{1}{c}{\small Acc} &  \multicolumn{1}{c}{\small Acc} & 
         \multicolumn{1}{c}{\small Acc}
           &  \multicolumn{1}{c}{\small Acc} 
 
         \\
    
    \hline\hline

         \small ViT-B/16 
         & \small \textit{davinci003} & \small \small \textbf{\color{gray}{73.6}}
         & \small \textbf{57.2}  
         & \small \textbf{63.8} 
         & \small \textbf{70.3} 
          & \small \textbf{64.0}  
         \\

         & \small \textit{gpt-3.5-turbo}
         & \small \small \color{gray}{71.9}  
         & \small \textbf{57.2}
         & \small 63.1
         & \small 69.9
          & \small 63.6
         \\
         \hline

         \small RN50  & \small \textit{davinci003} 
         & \small \textbf{\color{gray}{65.5}}
            & \small 48.1

         & \small \textbf{54.5} 
                 & \small 60.4
          & \small \textbf{54.8}  
         \\

         &  \small  \textit{gpt-3.5-turbo}
        & \small \small \color{gray}{64.8}
        & \small \small \textbf{48.3}  
        & \small 54.4 
        & \small \textbf{60.6}
        & \small \textbf{54.8}
          \\

    \hline
    \end{tabular}
    \end{center}
    \vspace{-5mm}
    \caption{\textbf{Regularizing soft prompts with geography knowledge, top-1 bal. accuracy on DollarStreet, \textit{davinci003} vs. \textit{gpt-3.5-turbo}.} Comparisons are shown for CountryInPrompt+LLM at $\lambda$=4. 
    While \textit{davinci003} is observably more performant with ViT-B/16, our strategy also works well with \textit{gpt-3.5-turbo}. }
    \label{table:supp_chatgpt}
\end{table}

        \noindent Table \ref{table:supp_chatgpt} shows our findings comparing \textit{gpt-3.5-turbo} vs. \textit{davinci-003} with geography knowledge regularization. \noindent \textit{davinci-003} is clearly the top LLM with respect to ViT-B/16, but results are more comparable with RN50. We leave a more rigid comparison study of LLM world knowledge for future work. 

        \noindent \textbf{Experiment: America as source.} We test America as a different source than Europe and show results in  Table \ref{table:rb3}. The results show improvements across target continents, exhibiting that our approach generalizes to other sources.

        \noindent \textbf{Experiment: Choice of ensemble.} Table \ref{table:rb1} shows results when varying the ensemble to be only from certain continents. Notably, having diverse continents (Am/Af/As) leads to top performance overall. A diverse ensemble may allow for a more domain-generalizable representation overall. Regarding single-continent performance, it is observed that the Am/As ensembles do not result in the top Am/As performance respectively. We believe that countries in other continents can be informative for learning (\eg if multiple countries use \textit{adobe} for houses). Future fine-grained analysis can be done to identify optimal ensemble properties.
        
         \setlength{\tabcolsep}{1.5pt}

\begin{table}[t]
    \begin{center}
        \renewcommand{\arraystretch}{0.8}

    \begin{tabular}{c|c|c|c|c|c}
        \hline
         &  \multicolumn{1}{c|}{\textit{\textbf{Source}}} 
         & \multicolumn{4}{c}{\textit{\textbf{Target}}}   \\
        
         Method & \multicolumn{1}{c|}{\textbf{Americas}} &\multicolumn{1}{c}{\textbf{Africa}} &\multicolumn{1}{c}{\textbf{Asia}} &\multicolumn{1}{c}{\textbf{Europe}} &\multicolumn{1}{c}{\textbf{Total}} \\
        
         &  \multicolumn{1}{c|}{\small Acc}  & 
         \multicolumn{1}{c}{\small Acc} &  \multicolumn{1}{c}{\small Acc} & 
         \multicolumn{1}{c}{\small Acc}
           &  \multicolumn{1}{c}{\small Acc} 
 
         \\
    
    \hline\hline

         \rowcolor{LGray} \small CoOp 
         & \small \color{gray}{71.1}
         & \small 56.4  
         & \small 62.5
         & \small 72.1
          & \small 64.2 
         \\

         \rowcolor{LGray} \small CoCoOp 
         & \small \small \color{gray}{70.7}  
         & \small 55.8 
         & \small 62.4
         & \small 72.2
          & \small 64.2 \\

         \rowcolor{LGray} \small KgCoOp 
         & \small \color{gray}{72.7}
       & \small 56.2
        & \small 63.0
         & \small 72.9
          & \small 64.0 \\

           \small CIPReg  
         & \small \small \color{gray}{71.4} \scriptsize \color{red}{-1.3} 
         & \small 57.4 \scriptsize \color{Greenish}{+1.0} 
         & \small 63.5 \scriptsize \color{Greenish}{+0.5} 
         & \small 72.6 \scriptsize \color{red}{-0.3} 
          & \small 64.6 \scriptsize \color{Greenish}{+0.4} 
         \\

         \small LLMReg 
         & \small \small \color{gray}{71.7} \scriptsize \color{red}{-1.0} 
         & \small \textbf{57.6} \scriptsize \color{Greenish}{+1.2} 
         & \small 63.8 \scriptsize \color{Greenish}{+0.8} 
         & \small 73.3 \scriptsize \color{Greenish}{+0.4} 
          & \small 64.8 \scriptsize \color{Greenish}{+0.6} \\

         \small CIP+LLMReg  
         & \small \textbf{\color{gray}{73.0}} \scriptsize \color{Greenish}{+0.3} 
       & \small 57.4 \scriptsize \color{Greenish}{+1.0} 
        & \small \textbf{64.0} \scriptsize \color{Greenish}{+1.0} 
         & \small \textbf{73.4} \scriptsize \color{Greenish}{+0.5} 
          & \small \textbf{65.1} \scriptsize \color{Greenish}{+0.9}

    \\

    \hline
    \end{tabular}
    \end{center}
    \vspace{-3mm}
    \caption{\textbf{Geo knowledge regularization on DollarStreet, src = Americas, tgt = Africa,Asia,Europe.} Encoder = ViT-B/16.}
    \label{table:rb3}
\end{table}

        \noindent \textbf{Experiment: Breaking down Americas.} Having observed drops for Americas/RN50 in Table 2 of the main paper, we breakdown overall accuracy for North America (USA, Canada, Mexico) and Central/South America (Haiti, Bolivia, Brazil, Colombia, Peru, Guatemala) in Table \ref{table:rb2}. Our method improves over KgCoOp in Central/South America for both encoders, but not in North America for RN50, which explains the overall drops. We reason that North America does not benefit from knowledge constraints due to CLIP already being well-aligned to images in countries like the USA.

         \setlength{\tabcolsep}{2.5pt}

\begin{table}[t]
    \begin{center}
        \renewcommand{\arraystretch}{0.9}

    \begin{tabular}{c|c|c|c|c|c|c}
        \hline
          & & \multicolumn{1}{c|}{\textit{\textbf{Source}}} 
         & \multicolumn{4}{c}{\textit{\textbf{Target}}}   \\
        
           \multicolumn{1}{c|}{\textbf{$\mathcal{G}_t$}}  & \multicolumn{1}{c|}{\textbf{n}} & \multicolumn{1}{c|}{\textbf{Eu}} &\multicolumn{1}{c}{\textbf{Af}} &\multicolumn{1}{c}{\textbf{As}} &\multicolumn{1}{c}{\textbf{Am}} &\multicolumn{1}{c}{\textbf{Total}} \\
        
         & & \multicolumn{1}{c|}{\small Acc}  & 
         \multicolumn{1}{c}{\small Acc} &  \multicolumn{1}{c}{\small Acc} & 
         \multicolumn{1}{c}{\small Acc}
           &  \multicolumn{1}{c}{\small Acc} 
 
         \\
    
    \hline\hline
    \small  
         \small Am/Af/As & 49 & \small \textbf{\color{gray}{73.6}} 
         
       & \small \textbf{57.2} 
        & \small \textbf{63.8} 
         & \small 70.3
          & \small \textbf{64.0}
          \\
          
         

         
      %
         \small Am & 9 & \small \color{gray}{73.4} 
       & \small 56.9
        & \small 63.6
         & \small 70.0
          & \small 63.7
          \\
           
        \small  Af & 18 & \small \color{gray}{73.2}
       & \small 57.0
        & \small 63.6
         & \small \textbf{70.4} 
          & \small 63.9

        \\
          
        \small  As & 22 & \small \color{gray}{73.4}
       & \small 57.0
        & \small 63.4  
         & \small 70.3
          & \small 63.6

    \\

    \hline
    \end{tabular}
    \end{center}
    \vspace{-3mm}
    \caption{\textbf{Varying geo. ensemble ($\mathcal{G}_t$) for CIP+LLMReg method, on DollarStreet.} Encoder=ViT-B/16. n=\# countries in ensemble.}
    \label{table:rb1}
\end{table}
         \setlength{\tabcolsep}{0.7pt}
\newcolumntype{g}{>{\columncolor{LGray}}c}

\begin{table}[t]
    \begin{center}
        \renewcommand{\arraystretch}{0.9}

    \begin{tabular}{c|c|g|c|g|c}
        \hline
                 &   &\multicolumn{2}{c|}{\textbf{ViT-B/16}} &\multicolumn{2}{c}{\textbf{RN50}} \\

         \small Region & \textbf{n} & \multicolumn{1}{c}{\textbf{\small KgCoOp}} &\multicolumn{1}{c|}{\textbf{\small CIP+LLMReg}} &\multicolumn{1}{c}{\small \textbf{KgCoOp}} &\multicolumn{1}{c}{\textbf{\small CIP+LLMReg}}  \\
        
         &    & 
         \multicolumn{1}{c}{\small Acc} &  \multicolumn{1}{c|}{\small Acc} & 
         \multicolumn{1}{c}{\small Acc}
           &  \multicolumn{1}{c}{\small Acc} 
 
         \\
    
    \hline\hline

         \small SA/CA 
         & \small 2,795
       & \small 66.6
        & \small \textbf{68.4} \scriptsize \color{Greenish}{+1.8}
         & \small 57.1
          & \small \textbf{57.9} \scriptsize \color{Greenish}{+0.8} \\

         \small NA 
         & \small 1,946  
       & \small 69.4 
        & \small \textbf{70.0} \scriptsize \color{Greenish}{+0.6} 
         & \small \textbf{60.9} 
          & \small 60.0 \scriptsize \color{red}{-0.9}

    \\

    \hline
    \end{tabular}
    \end{center}
    \vspace{-3mm}
    \caption{\textbf{Geo knowledge regularization on DollarStreet, perf. on North vs. South/Central Am. Src = Eu.} n = \# of test images. }
    \label{table:rb2}
\end{table}
         \setlength{\tabcolsep}{0.4pt}
   
\begin{table*}[t]
    \begin{center}
    \renewcommand{\arraystretch}{0.8}

    \begin{tabular}{c||cc|cc|cc|cc||cc|cc|cc|cc||cc|cc|cc|cc}

         \multicolumn{1}{c||}{} & \multicolumn{8}{c||}{\small \textbf{Africa}} & \multicolumn{8}{c||}{\small \textbf{Asia}} & \multicolumn{8}{c}{\small \textbf{Americas}}   \\
     \multicolumn{1}{c||}{} & \multicolumn{8}{c||}{\small \textbf{Threshold \textit{t} (\# Classes)}} & \multicolumn{8}{c||}{\small \textbf{Threshold \textit{t} (\# Classes)}} & \multicolumn{8}{c}{\small \textbf{Threshold \textit{t} (\# Classes)}}   \\
     \cline{2-25}

     \small \textbf{Method}
     & \multicolumn{2}{c|}{\small \textbf{$<$40\%}} 
     & \multicolumn{2}{c|}{\small \textbf{$<$60\%}} 
     & \multicolumn{2}{c|}{\small \textbf{$<$80\%}} 
     & \multicolumn{2}{c||}{\small \textbf{$\leq$100\%}} 
     & \multicolumn{2}{c|}{\small \textbf{$<$40\%}} 
     & \multicolumn{2}{c|}{\small \textbf{$<$60\%}} 
     & \multicolumn{2}{c|}{\small \textbf{$<$80\%}} 
     & \multicolumn{2}{c||}{\small \textbf{$\leq$100\%}}
     & \multicolumn{2}{c|}{\small \textbf{$<$40\%}} 
     & \multicolumn{2}{c|}{\small \textbf{$<$60\%}} 
     & \multicolumn{2}{c|}{\small \textbf{$<$80\%}} 
     & \multicolumn{2}{c}{\small \textbf{$\leq$100\%}}\\
     
     & (30) & \small $\Delta$ 
     & (55) & \small $\Delta$ 
     & (82) & \small $\Delta$ 
     & (94) & \small $\Delta$ 
     & (18) & \small $\Delta$ 
     & (42) & \small $\Delta$ 
     & (78) & \small $\Delta$ 
     & (95) & \small $\Delta$ 
     & (3) & \small $\Delta$ 
     & (30) & \small $\Delta$ 
     & (69) & \small $\Delta$ 
     & (92) & \small $\Delta$ \\
    \hline
    \hline
    \rowcolor{LLGray} \small CoOp \cite{zhou2022learning}
    & \small 27.4 & - 
    & \small 37.7 & -
    & \small 48.5 & - 
    & \small 53.9 & -
    & \small 30.9 & - 
    & \small 42.1 & -
    & \small 55.6 & - 
    & \small 61.5 & -
    & \small 24.9 & - 
    & \small 50.8 & - 
    & \small 61.9 & - 
    & \small 68.6 & -
\\

    \rowcolor{LLGray} \small CoCoOp \cite{Zhou_2022_CVPR}
    & \small 27.4 & \scriptsize \color{black}{0.0} 
    & \small 37.5 & \scriptsize \color{red}{-0.2} 
    & \small 48.7 & \scriptsize \color{Greenish}{+0.2} 
    & \small 54.3 & \scriptsize \color{Greenish}{+0.4} 
    & \small 32.6 & \scriptsize \color{Greenish}{+1.7} 
    & \small 42.3 & \scriptsize \color{Greenish}{+0.2} 
    & \small 55.4 & \scriptsize \color{red}{-0.2} 
    & \small 61.2 & \scriptsize \color{red}{-0.3} 
    & \small 33.1 & \scriptsize \color{Greenish}{+8.2} 
    & \small 51.9 & \scriptsize \color{Greenish}{+1.1} 
    & \small 61.5 & \scriptsize \color{red}{-0.4} 
    & \small 68.3 & \scriptsize \color{red}{-0.3} 
 \\
    
    \rowcolor{LLGray} \small KgCoOp  
    & \small 28.0 & \scriptsize \color{Greenish}{+0.6}  
    & \small 39.2 & \scriptsize \color{Greenish}{+1.5} 
    & \small 49.3 & \scriptsize \color{Greenish}{+0.8} 
    & \small 54.4 & \scriptsize \color{Greenish}{+0.5} 
    & \small 34.7 & \scriptsize \color{Greenish}{+3.8} 
    & \small 44.5 & \scriptsize \color{Greenish}{+2.4} 
    & \small 57.3 & \scriptsize \color{Greenish}{+1.7} 
    & \small 62.6 & \scriptsize \color{Greenish}{+1.1} 
    & \small 38.1 & \scriptsize \color{Greenish}{+13.2} 
    & \small 54.2 & \scriptsize \color{Greenish}{+3.4} 
    & \small 62.6 & \scriptsize \color{Greenish}{+0.7} 
    & \small 68.7 & \scriptsize \color{Greenish}{+0.1} 
\\
    
    \small CIPReg  
    & \small 31.9 & \scriptsize \color{Greenish}{+4.5}  
    & \small 41.5 & \scriptsize \color{Greenish}{+3.8} 
    & \small 51.7 & \scriptsize \color{Greenish}{+3.2} 
    & \small 56.8 & \scriptsize \color{Greenish}{+2.9} 
    & \small 35.9 & \scriptsize \color{Greenish}{+5.0} 
    & \small 45.1 & \scriptsize \color{Greenish}{+3.0} 
    & \small 57.8 & \scriptsize \color{Greenish}{+2.2} 
    & \small 63.0 & \scriptsize \color{Greenish}{+1.5} 
    & \small 40.7 & \scriptsize \color{Greenish}{+15.8} 
    & \small 55.6 & \scriptsize \color{Greenish}{+4.8} 
    & \small 63.5 & \scriptsize \color{Greenish}{+1.6} 
    & \small 69.8 & \scriptsize \color{Greenish}{+1.2} 
\\
    
    \small LLMReg  
    & \small 29.0 & \scriptsize \color{Greenish}{+1.6}  
    & \small 40.1 & \scriptsize \color{Greenish}{+2.4} 
    & \small 50.3 & \scriptsize \color{Greenish}{+1.8} 
    & \small 55.6 & \scriptsize \color{Greenish}{+1.7} 
    & \small 35.4 & \scriptsize \color{Greenish}{+4.5} 
    & \small 44.2 & \scriptsize \color{Greenish}{+2.1} 
    & \small 57.3 & \scriptsize \color{Greenish}{+1.7} 
    & \small 63.0 & \scriptsize \color{Greenish}{+1.5} 
    & \small 39.6 & \scriptsize \color{Greenish}{+14.7} 
    & \small 55.7 & \scriptsize \color{Greenish}{+4.9} 
    & \small 64.1 & \scriptsize \color{Greenish}{+2.2} 
    & \small 70.0 & \scriptsize \color{Greenish}{+1.4} 
 \\
    
    \small CIP+LLMReg  
    & \small \textbf{32.0} & \scriptsize \color{Greenish}{+4.6}  
    & \small \textbf{42.0} & \scriptsize \color{Greenish}{+4.3} 
    & \small \textbf{51.9} & \scriptsize \color{Greenish}{+3.4} 
    & \small \textbf{57.2} & \scriptsize \color{Greenish}{+3.3} 
    & \small \textbf{37.3} & \scriptsize \color{Greenish}{+6.4} 
    & \small \textbf{46.0} & \scriptsize \color{Greenish}{+3.9} 
    & \small \textbf{58.4} & \scriptsize \color{Greenish}{+2.8} 
    & \small \textbf{63.8} & \scriptsize \color{Greenish}{+2.3} 
    & \small \textbf{46.5} & \scriptsize \color{Greenish}{+21.6} 
    & \small \textbf{56.8} & \scriptsize \color{Greenish}{+6.0} 
    & \small \textbf{64.4} & \scriptsize \color{Greenish}{+2.5} 
    & \small \textbf{70.3} & \scriptsize \color{Greenish}{+1.7} 
 \\
    \hline
    \end{tabular}
    \end{center} 
    \vspace{-3mm}
        \caption{\textbf{Performance on DollarStreet classes with less than \textit{t}\% recall in CoOp, respective to continents}, with ViT-B/16. Shown are \color{Greenish}{gains}\color{black}/\color{red}{drops }\color{black} w.r.t. CoOp. Our top method improves greatest in the $<$40\% scenario, in \textit{every} continent scenario. CIP = CountryInPrompt, LLM = CountryLLM, CIP+LLM= CountryInPrompt+LLM.}
    \label{table:dollarstreet_cont_per_class_perf}
\end{table*}

         \setlength{\tabcolsep}{0.4pt}
   
\begin{table}[!tp]
    \begin{center}
    \renewcommand{\arraystretch}{0.8}

    \begin{tabular}{c|cc|cc|cc|cc|cc}
     \multicolumn{1}{c|}{} & \multicolumn{10}{c}{\small \textbf{Threshold \textit{t} (\# Classes)}} \\
     \cline{2-11}

     \small \textbf{Method}
     & \multicolumn{2}{c|}{\small \textbf{$<$5\%}}
     & \multicolumn{2}{c|}{\small \textbf{$<$40\%}} 
     & \multicolumn{2}{c|}{\small \textbf{$<$60\%}} 
     & \multicolumn{2}{c|}{\small \textbf{$<$80\%}} 
     & \multicolumn{2}{c}{\small \textbf{$\leq$100\%}}  \\
     
     & (36) & \small $\Delta$ 
     & (212) & \small $\Delta$ 
     & (345) & \small $\Delta$ 
     & (483) & \small $\Delta$ 
     & (600) & \small $\Delta$  \\
    \hline
    \hline
    \rowcolor{LLGray} \small CoOp \cite{zhou2022learning}
    & \small 1.4 & - 
    & \small 19.6 & - 
    & \small 30.7 & - 
    & \small 41.9 & - 
    & \small 51.2 & -\\

    \rowcolor{LLGray} \small CoCoOp \cite{Zhou_2022_CVPR}
    & \small 4.4 & \scriptsize \color{Greenish}{+3.0} 
    & \small 22.2 & \scriptsize \color{Greenish}{+2.6} 
    & \small 33.5 & \scriptsize \color{Greenish}{+2.8} 
    & \small 43.7 & \scriptsize \color{Greenish}{+1.8} 
    & \small 52.6 & \scriptsize \color{Greenish}{+1.4}\\
    
    \rowcolor{LLGray} \small KgCoOp  
    \cite{yao2023visual} 
    & \small 4.3 & \scriptsize \scriptsize \color{Greenish}{+2.9}
    & \small 22.4 & \scriptsize \scriptsize \color{Greenish}{+2.8}
    & \small 34.3 & \scriptsize \scriptsize \color{Greenish}{+3.6} 
    & \small 43.9 & \scriptsize \scriptsize \color{Greenish}{+2.0} & \small 52.6 & \scriptsize \scriptsize \color{Greenish}{+1.4} \\
    
    \small CIPReg  
    & \small 6.7 & \scriptsize \color{Greenish}{+5.3} 
    & \small\textbf{24.2} & \scriptsize \color{Greenish}{+4.6} 
    & \small 35.5 & \scriptsize \color{Greenish}{+4.8} 
    & \small 45.0 & \scriptsize \color{Greenish}{+3.1}
    & \small 53.5 & \scriptsize \color{Greenish}{+2.3}\\
    
    \small LLMReg   
    & \small 3.6 & \scriptsize \color{Greenish}{+2.2} 
    & \small 23.2 & \scriptsize \color{Greenish}{+3.6} 
    & \small 34.8 & \scriptsize \color{Greenish}{+4.1} 
    & \small 44.5 & \scriptsize \color{Greenish}{+2.6} & \small 53.1 & \scriptsize \color{Greenish}{+1.9}\\
    
    \small CIP+LLMReg  
    & \small \textbf{7.0} & \scriptsize \color{Greenish}{+5.6} 
    & \small \textbf{24.2} & \scriptsize \color{Greenish}{+4.6} 
    & \small \textbf{35.8} & \scriptsize \color{Greenish}{+5.1} 
    & \small \textbf{45.4} & \scriptsize \color{Greenish}{+3.5}
    & \small \textbf{53.9} & \scriptsize \color{Greenish}{+2.7} \\
    \hline
    \end{tabular}
    \end{center} 
    \vspace{-3mm}
        \caption{\textbf{Performance on GeoNet classes with less than \textit{t}\% recall in CoOp}, with ViT-B/16. Gains w.r.t. CoOp of geography knowledge regularization are especially large for CoOp's difficult classes (+5.6 in $<$5\%). CIP = CountryInPrompt, LLM = CountryLLM, CIP+LLM= CountryInPrompt+LLM.}
    \label{table:supp_geonet_per_class_perf}
\end{table}
        \begin{figure*}[!h]
    \centering
    \begin{subfigure}{0.49\textwidth}
        \includegraphics[width=\textwidth]{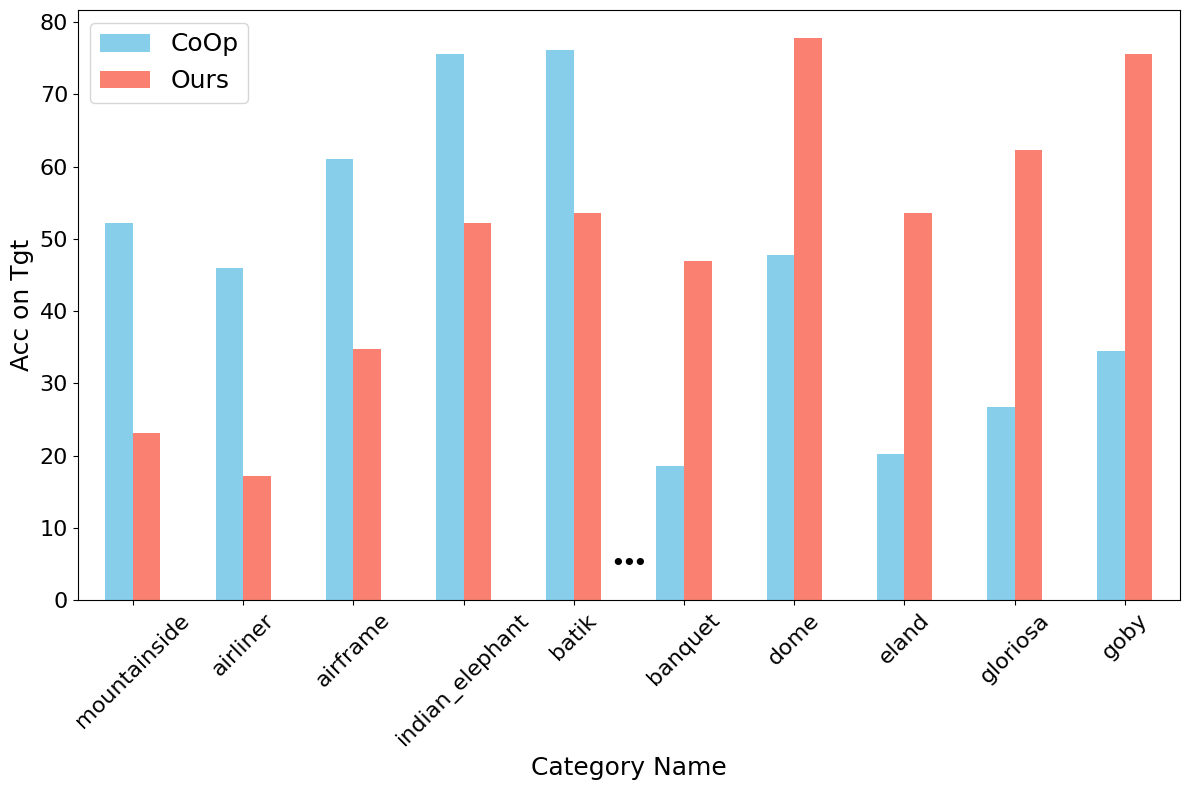}
        \caption{\textbf{GeoNet}}
        \label{supp:histogram_geonet}
    \end{subfigure}
    \hfill 
    \begin{subfigure}{0.49\textwidth}
        \includegraphics[width=\textwidth]{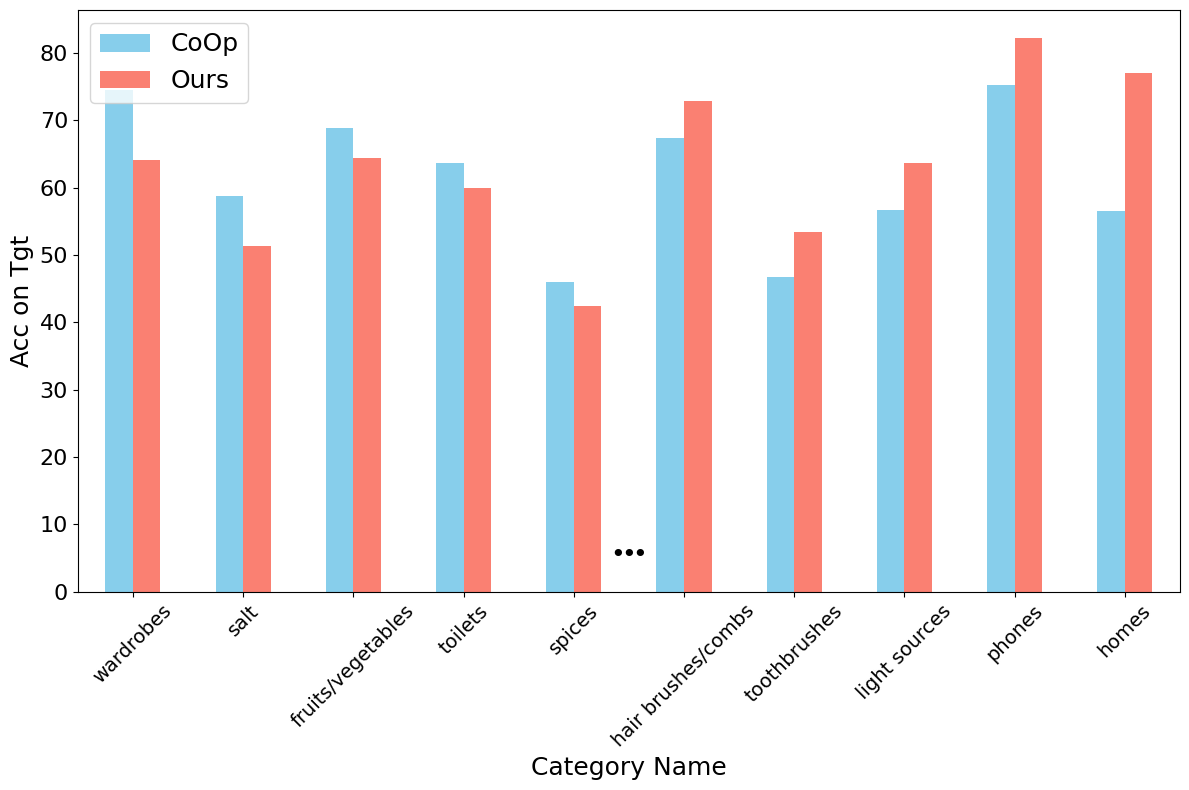}
        \caption{\textbf{DollarStreet}}
        \label{supp:histogram_dollarstreet}
    \end{subfigure}
    \vspace{-3mm}
    \caption{\textbf{Classwise comparison of Ours (CountryInPrompt+LLM) vs. CoOp on GeoNet and Dollarstreet.} We sort the top 25 most frequent categories based on $\Delta = Acc(Ours) - Acc(CoOp)$ and show the bottom 5 and top 5 categories. Leveraging geographical knowledge can be most useful on objects that may have geography-specific characteristics, while it may hurt generic category performance.}
    \label{supp:histogram_combined}
\end{figure*}

        \noindent \textbf{Classes by difficulty by continent: DollarStreet.} In the main paper (Table 4), we show performance on ``difficult'' classes for CoOp overall. We provide further results analyzing performance on difficult classes with respect to each continent in DollarStreet, shown in Table \ref{table:dollarstreet_cont_per_class_perf}. Similarly, our top method provides the top gains on the difficult classes, i.e. the $<$40\% scenario, across every continent. 
        
        \noindent \textbf{Classes with the most impact:  GeoNet \& DollarStreet.} In Figure~\ref{supp:histogram_combined}, we show classes where our regularization method has the most impact, in both the positive direction (gains) and negative direction (drops). On DollarStreet, it is notably effective for \textit{homes}, which vary in appearance and construction materials across regions. On GeoNet, it benefits categories like \textit{goby} (a type of fish), \textit{gloriosa} (a type of flower), and \textit{dome}, with domes differing in color between the USA (typically gray) and Asian countries (often yellow and orange). While \textit{goby} and \textit{gloriosa} generally look consistent worldwide, their images may experience context shifts due to environmental differences. On the other hand, general categories such as \textit{airliner}, \textit{mountainside}, and \textit{salt} are adversely affected by geographical knowledge regularization. Ensuring good performance across all classes, perhaps through considering the adaptation of class representations at a finer-grained level, is needed in future work. 

        \noindent \textbf{Classes by difficulty: GeoNet.} We show a per-class breakdown of our knowledge regularization on GeoNet in Table \ref{table:supp_geonet_per_class_perf}. Like with DollarStreet in Table 4 of main, we show the thresholds $t$=40,60,80,100; however, since GeoNet has a large number of classes, we also show $t$=5 for a more aggressive threshold. We observe a similar observation that our method performs well on the most challenging classes at $t$=5 (+5.6\% vs. CoOp baseline). CountryInPrompt regularization appears to drive performance of the hard classes in this case; CountryLLM regularization provides more evenly distributed improvements across thresholds. 
        
        

        \begin{table}[t]
\label{cvd}
    \begin{center}
        \renewcommand{\arraystretch}{1}
        \begin{tabular}{|c|c|c|c|c|c|c|c|c|c|}
            \hline
                \small \textbf{Class} & \small \textbf{Descriptor} & \multicolumn{2}{c|}{\textbf{Eu}} 
                 & \multicolumn{2}{c|}{\textbf{Af}} 
                 & \multicolumn{2}{c|}{\textbf{As}} 
                 & \multicolumn{2}{c|}{\textbf{Am}}   \\
                 \cline{3-10}
                 & & \small Count & \small Rel & \small Count & \small Rel & \small Count & \small Rel & \small Count & \small Rel \\
                 
                \hline\hline 
           
                 \small home & \small stone & \small 7 & \small 0.47 & \small 2 & \small 0.11 & \small 5 & \small 0.24 & \small 2 & \small 0.22 \\
                \small & \small mud  & \small 0 & \small 0.00 & \small 16 & \small 0.89 & \small 6 & \small 0.29 & \small 0 & \small 0.00 \\
                \small & \small bright colors & \small 2 & \small 0.13 & \small 12 & \small 0.67 & \small 8 & \small 0.38 & \small 2 & \small 0.22\\
                \small & \small balcony & \small 14 & \small 0.93 & \small 2 & \small 0.11 & \small 14 & \small 0.67 & \small 8 & \small 0.89\\
                \small & \small flower & \small 7 & \small 0.47 & \small 0 & \small 0.00 & \small 0 & \small 0.00 & \small 3 & \small 0.33 \\
                \hline

                \small toilet & \small squat & \small 0 & \small 0.00 & \small 6 & \small 0.33 & \small 4 & \small 0.19 & \small 1 & \small 0.11\\
                \small & \small white & \small 15 & \small 1.00 & \small 14 & \small 0.78 & \small 20 & \small 0.95 & \small 9 & \small 1.00\\
                \small & \small bidet & \small 7 & \small 0.47 & \small 1 & \small 0.06 & \small 9 & \small 0.43 & \small 3 & \small 0.33\\
                \small & \small button & \small 6 & \small 0.40 & \small 2 & \small 0.11 & \small 5 & \small 0.24 & \small 1 & \small 0.11 \\
                \small & \small ceramic & \small 8 & \small 0.53 & \small 12 & \small 0.67 & \small 14 & \small 0.67 & \small 5 & \small 0.56\\
                
                \hline
                 \small roof & \small thatch & \small 1 & \small 0.07 & \small 9  & \small 0.50 & \small 10 & \small 0.48 & \small 3 & \small 0.33 \\
                \small & \small straw & \small 0 & \small 0.00 & \small 12 & \small 0.67 & \small 4 & \small 0.19 & \small 0 & \small 0.00\\
                \small & \small terracotta & \small 4 & \small 0.27 & \small 1 & \small 0.06 & \small 4 & \small 0.19 & \small 3 & \small 0.33\\
                \small & \small metal & \small 4 & \small 0.27 & \small 8 & \small 0.44 & \small 9 & \small 0.43 & \small 6 & \small 0.67 \\
                \small & \small clay & \small 3 & \small 0.20 & \small 8 & \small 0.44 & \small 8 & \small 0.38 & \small 5 & \small 0.56 \\
                \hline
        \end{tabular}
    \end{center}
    \vspace{-3mm}
    \caption{\textbf{Example descriptor topics.} For various DollarStreet classes, we show examples of common words in the LLM descriptor sets across countries (grouped by continent). Count is the overall frequency within a continent, while rel. is the relative count (normalized by amount of countries in continent in DollarStreet). Country counts: Eu 15, Af 18, As 21, Am 9.}
    \label{tab:supp_desc_topics}

\end{table}
        \begin{figure*}[b]
    \begin{minipage}[c]{0.1\linewidth}
        \centering
        \vspace{0pt} 
        Home
    \end{minipage}%
    \begin{subfigure}[c]{0.44\linewidth}
        \includegraphics[height=0.4\linewidth, width=1\linewidth]{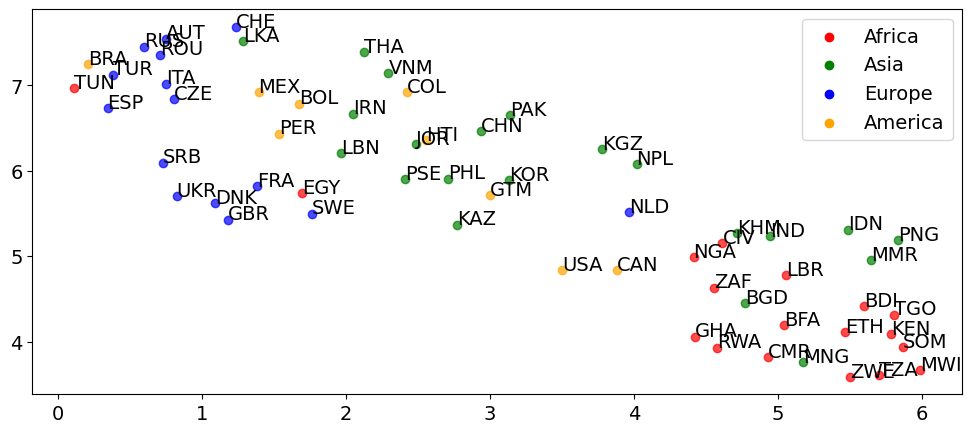}
    \end{subfigure}
    \begin{subfigure}[c]{0.44\linewidth}
        \includegraphics[height=0.4\linewidth, width=1\linewidth]{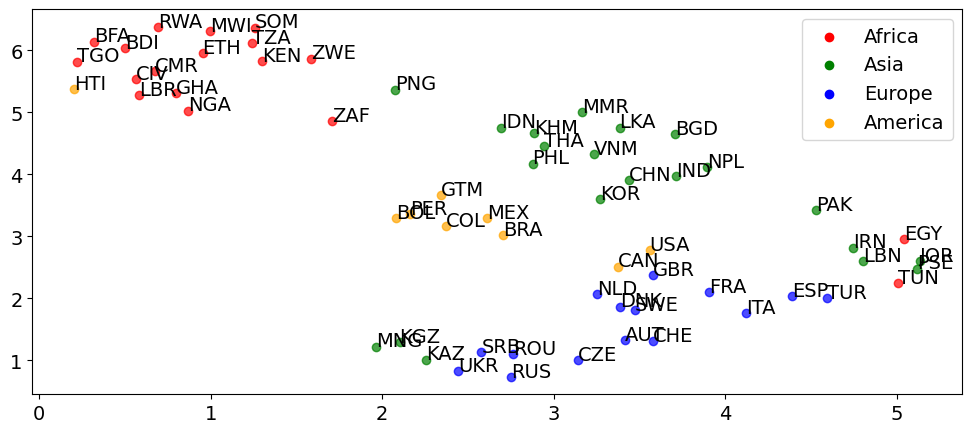}
    \end{subfigure}

    
    \begin{minipage}[c]{0.094\linewidth}
        \centering
        \vspace{0pt} 
        Jewelry
    \end{minipage}%
    \begin{subfigure}[c]{0.44\linewidth}
        \includegraphics[height=0.4\linewidth, width=1\linewidth]{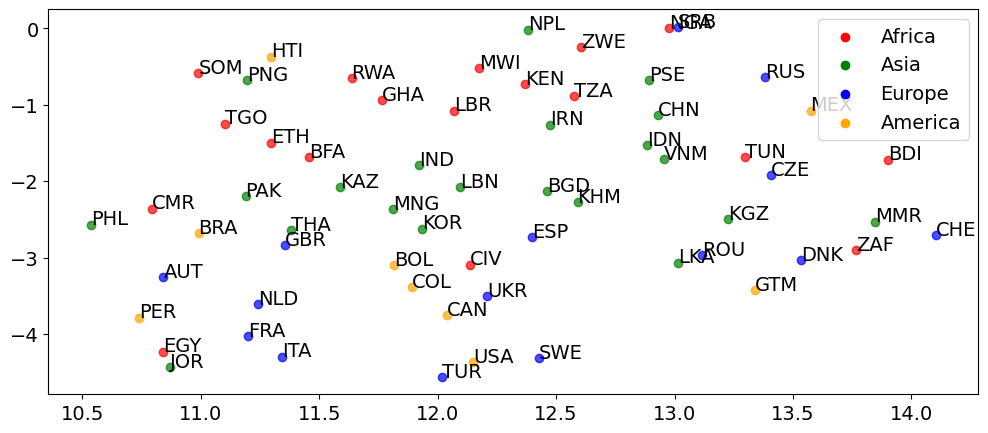}
    \end{subfigure}
    \begin{subfigure}[c]{0.44\linewidth}
        \includegraphics[height=0.4\linewidth, width=1\linewidth]{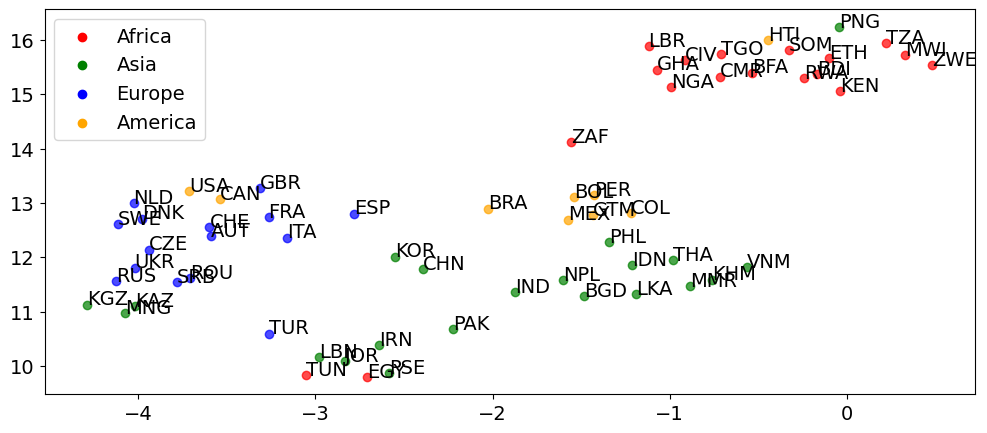}
    \end{subfigure}

    \begin{minipage}[c]{0.1\linewidth}
        \centering
        \vspace{0pt} 
        Ovens
    \end{minipage}%
    \begin{subfigure}[c]{0.44\linewidth}
        \includegraphics[height=0.4\linewidth, width=1\linewidth]{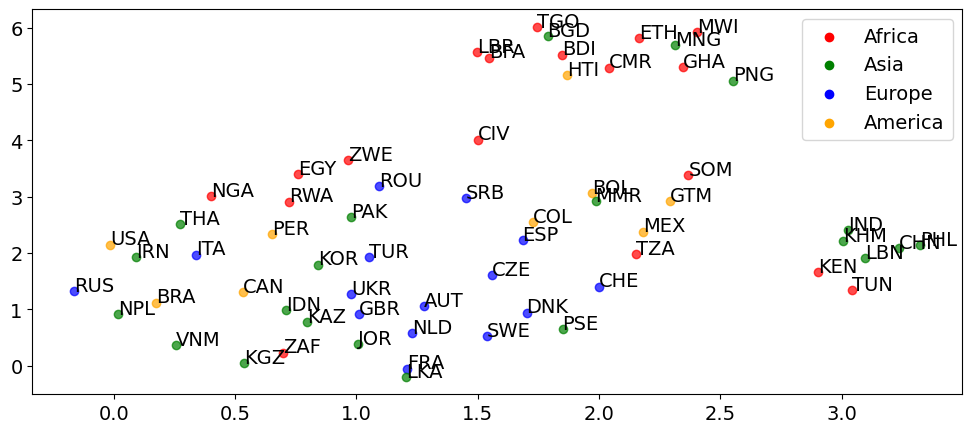}
    \end{subfigure}
    \begin{subfigure}[c]{0.44\linewidth}
        \includegraphics[height=0.4\linewidth, width=1\linewidth]{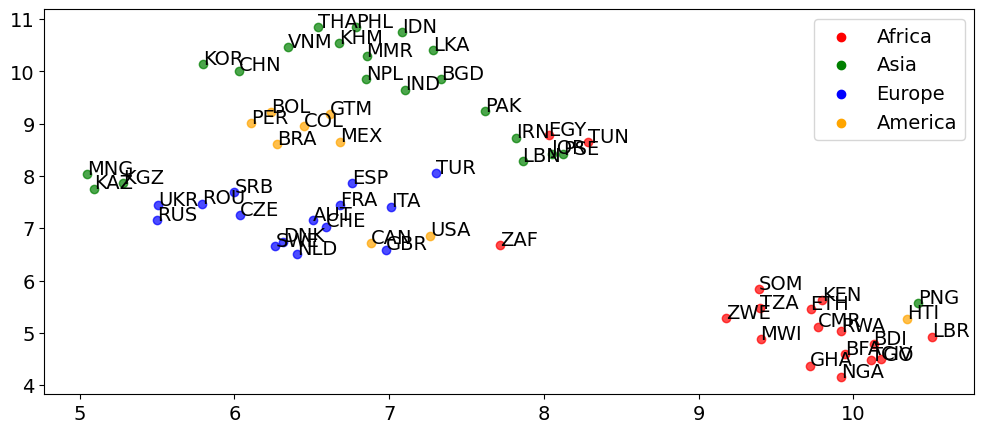}
    \end{subfigure}

     \begin{minipage}[c]{0.1\linewidth}
        \centering
        \vspace{0pt} 
        Roofs
    \end{minipage}%
    \begin{subfigure}[c]{0.44\linewidth}
        \includegraphics[height=0.4\linewidth, width=1\linewidth]{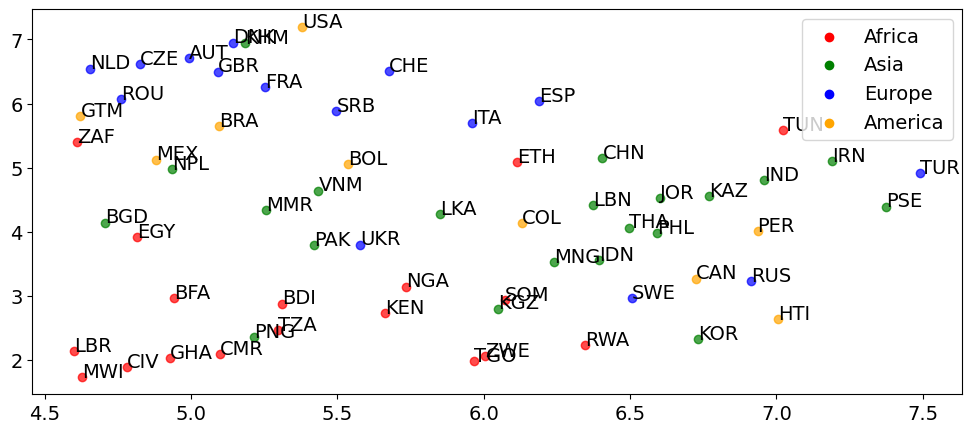}
    \end{subfigure}
    \begin{subfigure}[c]{0.44\linewidth}
        \includegraphics[height=0.4\linewidth, width=1\linewidth]{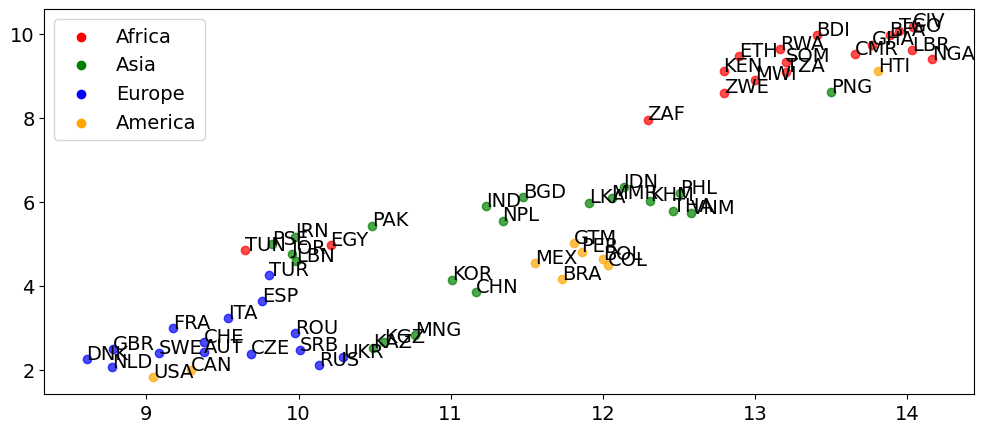}
    \end{subfigure}

    \begin{minipage}[c]{0.1\linewidth}
        \centering
        \vspace{0pt} 
        Toilets
    \end{minipage}%
    \begin{subfigure}[c]{0.44\linewidth}
        \includegraphics[height=0.4\linewidth, width=1\linewidth]{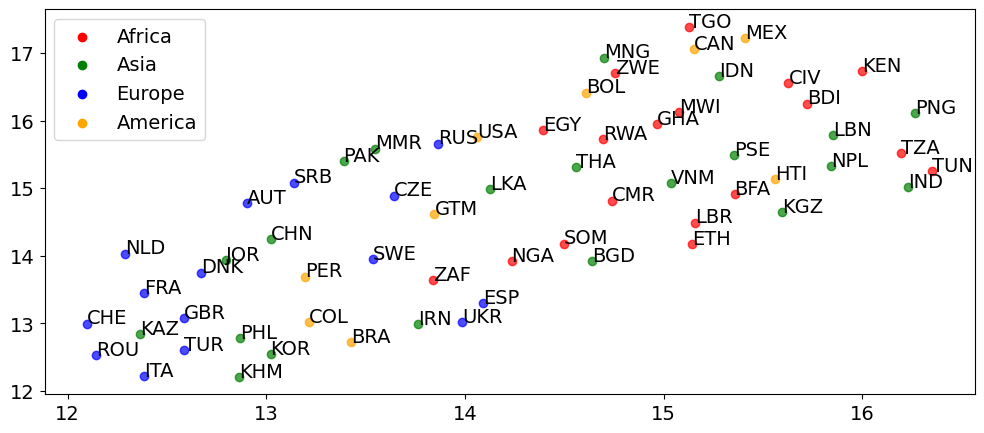}
    \end{subfigure}
    \begin{subfigure}[c]{0.44\linewidth}
        \includegraphics[height=0.4\linewidth, width=1\linewidth]{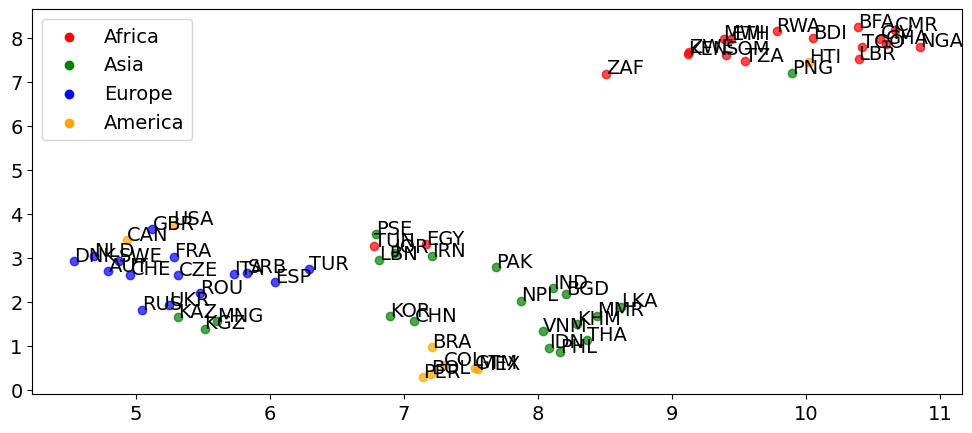}
    \end{subfigure}


    \begin{minipage}[c]{0.1\linewidth}
        \centering
        \vspace{0pt} 
        Toys
    \end{minipage}%
    \begin{subfigure}[c]{0.44\linewidth}
        \includegraphics[height=0.4\linewidth, width=1\linewidth]{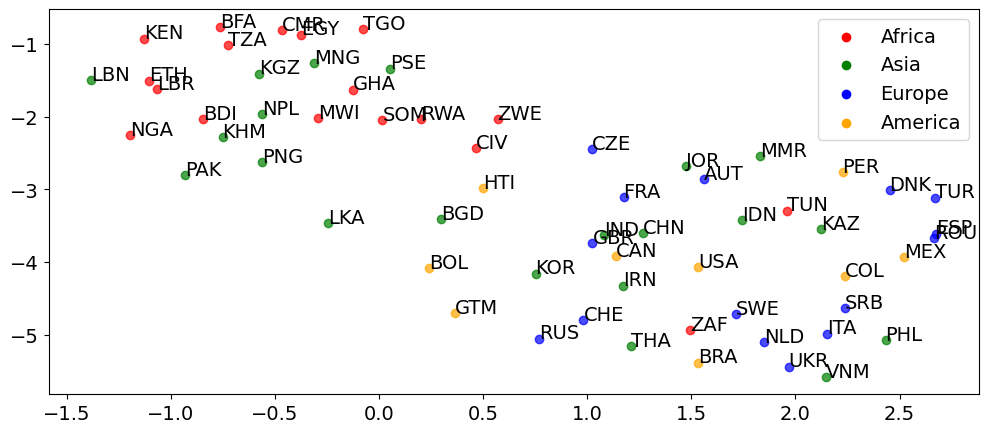}
        \caption{CountryLLM}
    \end{subfigure}
    \begin{subfigure}[c]{0.44\linewidth}
        \includegraphics[height=0.4\linewidth, width=1\linewidth]{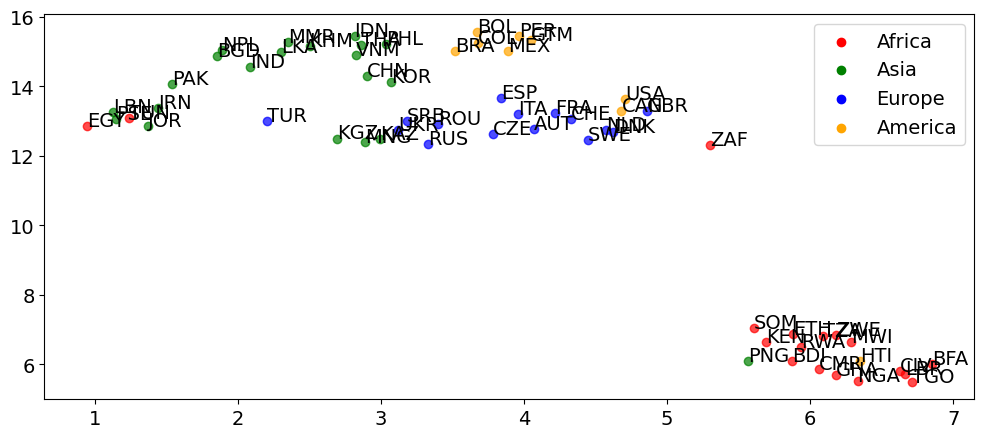}
        \caption{CountryInPrompt + LLM}
    \end{subfigure}
    \vspace{-3mm}
    \caption{\textbf{UMAP plots for various DollarStreet categories, CountryLLM vs. CountryInPrompt+LLM}. Country-specific class text embeddings often are close to those of neighboring countries. When CLIP's internal knowledge is added from (a) to (b) with the addition of country names, the clusters tighten.}
    \label{supp:umap}
\end{figure*}

    \subsection{Further Analysis}

        \noindent \textbf{Descriptor topics.} In Table \ref{tab:supp_desc_topics}, we show examples of words that appear amongst the geography-specific LLM descriptors for various DollarStreet categories. There exist some significant differences. For instance, for \textit{toilet}, ``squat'' appears multiple times in Africa and Asia descriptors, but not in European descriptors. Similarly, for \textit{roof}, ``thatch'' is more common in Africa and Asia than Europe and Americas. There are also some notable concepts that are common across regions, such as a \textit{toilet} being ``white'' and \textit{roof} being ``metal''. We advocate for future work that ensures factuality and proper representativeness of such concepts to extend utility to various regions. 

        \noindent \textbf{UMAP for further categories.} In Figure~\ref{supp:umap}, UMAP  \cite{mcinnes2018umap-software} visualization is used to compare the class text embeddings of CountryLLM and CountryInPrompt+LLM across various DollarStreet classes. With CountryLLM, we note that LLM descriptors are often more alike among countries within the same continent than between different continents. For instance, due to cultural differences, people may use different kinds of toilets in Africa compared to European countries. CountryInPrompt+LLM on the other hand shows much tighter clusters of countries, especially intra-continent, due to the addition of CLIP's internal knowledge.

\end{document}